%% file: output.tex
\documentclass{article} 

\usepackage[preprint]{colm2026_conference}

\usepackage{microtype}
\usepackage{hyperref}
\usepackage{url}
\usepackage{booktabs}

\usepackage{lineno}

\usepackage{microtype}
\usepackage{graphicx}

\usepackage{subfigure}
\usepackage{subcaption}
\usepackage{enumitem}
\usepackage{booktabs} 
\usepackage{multirow}
\usepackage[table,xcdraw]{xcolor} 
\usepackage{caption}
\usepackage{hyperref}
\usepackage{xspace}
\usepackage{fontawesome}   
\usepackage{wrapfig}  
\usepackage{multirow, booktabs, colortbl}
\usepackage{algorithm2e}
\usepackage{algorithmic}

\newcommand{\model}{\texttt{RationalRewards}\xspace}
\newcommand{\framework}{PARROT\xspace}

\input{math_commands.tex}

\usepackage{hyperref}
\usepackage{url}

\usepackage{amsmath}
\usepackage{amssymb}
\usepackage{mathtools}
\usepackage{amsthm}
\usepackage[most]{tcolorbox}
\tcbuselibrary{breakable}
\tcbuselibrary{listings}
\tcbuselibrary{skins} 
\usepackage[capitalize,noabbrev]{cleveref}

\theoremstyle{plain}

\theoremstyle{definition}

\theoremstyle{remark}

\usepackage[textsize=tiny]{todonotes}
\let\oldtodo\todo
\renewcommand{\todo}[1]{\oldtodo[color=red!35]{#1}}

\newlist{tightitem}{itemize}{1}
\setlist[tightitem]{label=--, leftmargin=1.2em, itemsep=1pt, topsep=2pt}
\newlist{tightenum}{enumerate}{1}
\setlist[tightenum]{label=\arabic*., leftmargin=1.4em, itemsep=1pt, topsep=2pt}

\definecolor{darkblue}{rgb}{0, 0, 0.5}
\hypersetup{colorlinks=true, citecolor=darkblue, linkcolor=darkblue, urlcolor=darkblue}

\title{RationalRewards: Reasoning Rewards Scale Visual Generation Both Training and Test Time}

\author{
\hspace{-5pt}\textbf{Haozhe Wang}\textsuperscript{1} \quad
\hspace{-5pt}\textbf{Cong Wei}\textsuperscript{2} \quad
\hspace{-5pt}\textbf{Weiming Ren}\textsuperscript{2} \quad
\hspace{-5pt}\textbf{Jiaming Liu}\textsuperscript{3} \quad
\hspace{-5pt}\textbf{Fangzhen Lin}\textsuperscript{1} \quad
\hspace{-5pt}\textbf{Wenhu Chen}\textsuperscript{2}\\[0.5em]
\textsuperscript{1}\,HKUST \quad
\textsuperscript{2}\,University of Waterloo\quad
\textsuperscript{3}\,Alibaba 
}

\begin{document}
\include{figures_def.tex}
\include{prompt_template.tex}

\ifcolmsubmission
\linenumbers
\fi

\maketitle

\begin{abstract}
    Most reward models for visual generation reduce rich human judgments to a single unexplained score, discarding the reasoning that underlies preference. We show that teaching reward models to produce explicit, multi-dimensional critiques before scoring transforms them from passive evaluators into active optimization tools—improving generators in two complementary ways: at training time, structured rationales provide interpretable, fine-grained rewards for reinforcement learning; at test time, a Generate–Critique–Refine loop turns critiques into targeted prompt revisions that improve outputs without any parameter updates. To train such a model without costly rationale annotations, we introduce Preference-Anchored Rationalization (PARROT), a principled framework that recovers high-quality rationales from readily available preference data through anchored generation, consistency filtering, and distillation. The resulting model, \textbf{RationalRewards} (8B), achieves state-of-the-art preference prediction among open-source reward models—competitive with Gemini-2.5-Pro—while using 10–20× less training data than comparable baselines. As an RL reward, it consistently improves text-to-image and image-editing generators beyond scalar alternatives. Most strikingly, its test-time critique-and-refine loop matches or exceeds RL-based fine-tuning on several benchmarks, suggesting that structured reasoning can unlock latent capabilities in existing generators that suboptimal prompts fail to elicit. 
Models and Code are available at \faGlobe\ 
\href{https://tiger-ai-lab.github.io/RationalRewards/}{{\textbf{Project Page}}}.
    
    \end{abstract}

\vspace{-2em}
\begin{figure}[h!]
    \centering
    
    \includegraphics[width=\textwidth]{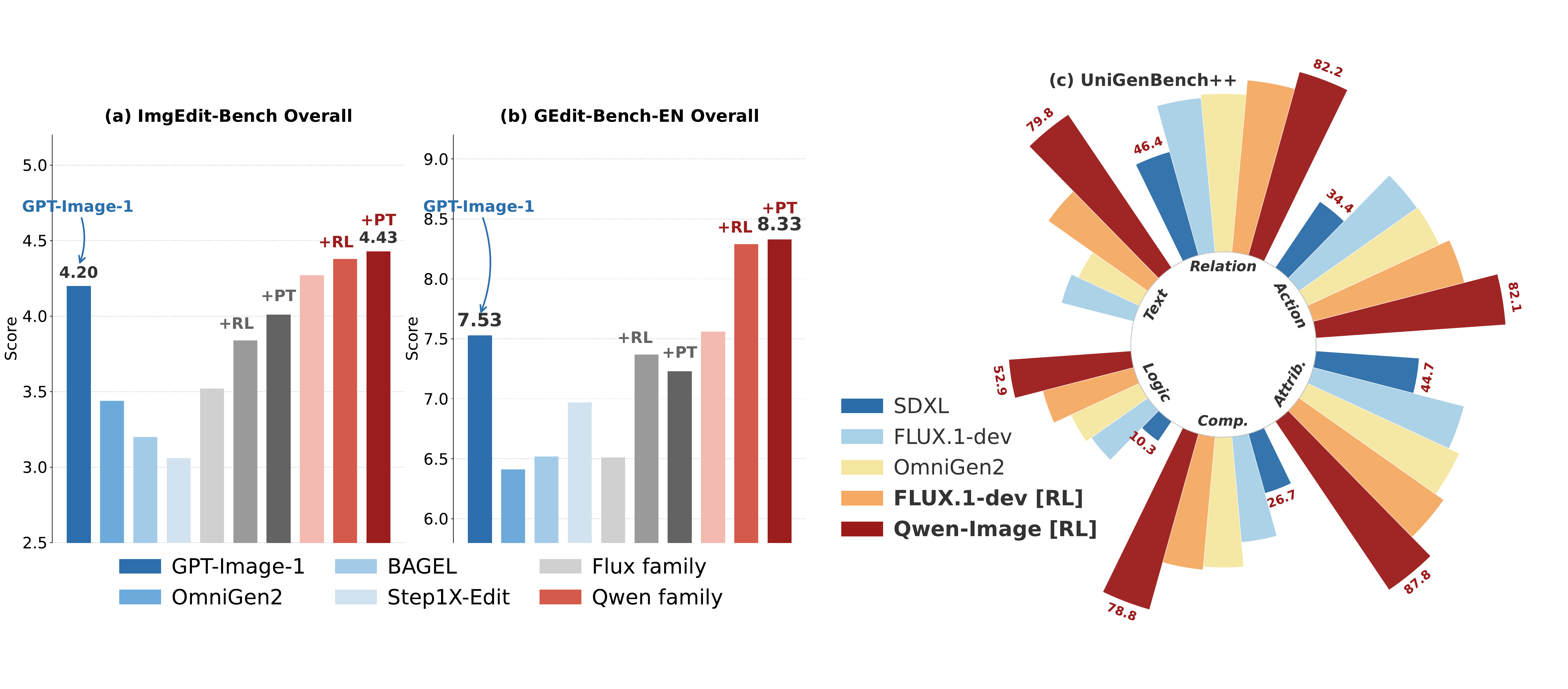}
    \vspace{-1cm}
    \caption{\small Train-Time RL and Test-Time PromptTuning (PT) with \model on text and image-to-image generation benchmarks. (Left) Comparison on image editing benchmarks. RL with \model outperforms prior open-source generators. Crucially, we find that test-time PT  with \model alone can surpass expensive RL. (Right) Breakdown results on text-to-image benchmark UniGenBench++.}
    \label{fig:performance_teaser}
    \vspace{-.3cm}
\end{figure}

\section{introduction}
As visual generation advances toward photorealistic, instruction-following outputs~\citep{nanobanana, gptimage, qwenimage, esser2024scaling}, reward models that \textit{evaluate} these outputs have become the binding constraint on further progress. Yet most reward models remain scalar black boxes: they compress multi-dimensional human judgments---perceptual quality, instruction faithfulness, physical plausibility, text rendering---into a single unexplained number~\citep{imagereward, editreward, liu2025improving, wei2024omniedit, mmrb2}. This discards the structured reasoning underlying human preference, leaving generators to exploit shortcut correlations rather than learn principled evaluation criteria~\citep{editr1}. This paper asks: can reward models be made to \textit{reason}---and can their structured critiques not only evaluate but actively \textit{improve} visual generation? 

\usagefigure
We introduce \model, a reasoning-based reward model that 
generates structured, multi-dimensional critiques before deriving scores. 
We argue that this shift from scalar outputs to structured reasoning transforms the reward model from a passive evaluator into a \textbf{versatile optimization interface} for 
visual generation. By producing explicit reasoning, \model unlocks 
optimization in two complementary spaces:

\begin{itemize}[leftmargin=*,itemsep=0pt, topsep=0pt]
\item \textbf{Parameter Space:} Multi-dimensional structured rationales provide semantically grounded, dense feedback for reinforcement learning---replacing opaque scalar gradients prone to reward hacking (Fig.~\ref{fig_reward_hacking_example}), with explanations of \textit{what} to improve and \textit{why}.
\item \textbf{Prompt Space:} Beyond serving as a reward signal, \model functions as a \textit{post-generation prompt optimizer}. It critiques a generated image, identifies concrete deficiencies, and 
translates them into targeted prompt revisions in a 
\textbf{Generate--Critique--Refine} loop. Unlike prompt enhancers that 
rewrite inputs blindly before synthesis~\citep{promptenhancer}, 
this approach is post-hoc and reactive, trading test-time compute for 
improved fidelity without parameter 
updates~\citep{snell2024scaling,wang2025vl}.
\end{itemize}

\def\methodfigure{
\begin{figure}[t!]
    \centering
    \vspace{-.8cm}\includegraphics[width=\linewidth]{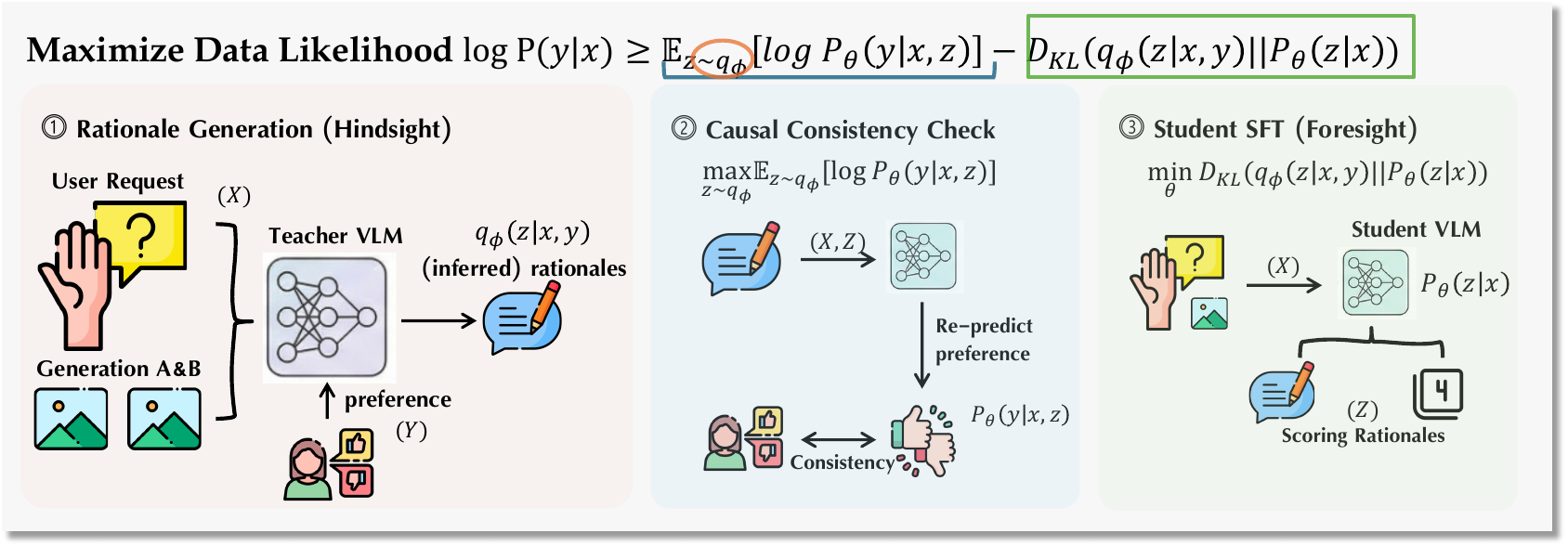}
    \vspace{-.4cm}
    \caption{\small \textbf{We implement Preference-Anchored Rationalization as a practical three-phase pipeline.} }
    \vspace{-.4cm}
    \label{fig_PARROT}
\end{figure}
}

Realizing this vision requires a reward model that produces high-quality 
structured rationales~\citep{genrm, rrm, star, reer}, yet human rationale annotations are prohibitively 
expensive at scale. We observe, however, that pairwise preference data is 
widely available from online AIGC platforms. Leveraging this, we propose 
\textbf{Preference-Anchored Rationalization (PARROT)}, a variational training framework 
that treats rationales as latent variables and derives an evidence lower 
bound (ELBO) on observed preferences. The terms of this ELBO map directly onto a simple, scalable pipeline: (1)~a teacher VLM generates candidate rationales anchored to known preference labels, (2)~a consistency filter rejects hallucinations and retains rationales that are genuinely predictive, and (3)~a student model is trained to produce rationales without seeing the answer. This tight theory--practice correspondence (Fig.~\ref{fig_PARROT}) converts existing preference datasets into high-quality reasoning supervision using 10--20$\times$ less data than comparable scalar reward baselines. 

\rewardhackfigure
\textbf{Key results.} Instantiated via PARROT on Qwen3-VL-Instruct-8B backbone, \model achieves state-of-the-art preference prediction among open-source reward models, competitive with 
Gemini-2.5-Pro (Table~\ref{tab:main_results}). As an RL reward, it consistently improves generators beyond scalar baselines across both text-to-image and image editing tasks (Tables~\ref{tab:unigen_ablation}--\ref{tab:ablation_sidebyside}). Most interestingly, \model's Generate--Critique--Refine loop---requiring no parameter updates---matches or exceeds RL-based fine-tuning on several benchmarks, suggesting that structured critiques can unlock latent generator capabilities that suboptimal prompts fail to elicit. We envision that \model empower more than four compelling use cases demonstrated in Fig.~\ref{fig_usage}.

\section{Method}
\label{sec:method}

We introduce \textbf{Preference-Anchored Rationalization (PARROT)}, a framework that trains reward models to produce explicit, multi-dimensional rationales before scores~\citep{star,reer}. Assessment dimensions---text faithfulness, physical/visual quality, text rendering, and (for editing) image faithfulness---follow the taxonomy of~\citet{mmrb2}, chosen for coverage of the primary failure modes in current generators.

Since ground-truth rationales are prohibitively expensive to annotate at scale, we formulate rationales as \textit{latent variables} inferred from pairwise preference data via a variational objective. The resulting ELBO (Eq.~\ref{eq_distill}) decomposes into three terms, each corresponding to a concrete pipeline phase (Fig.~\ref{fig_PARROT}): (1)~generate rationales anchored to known preferences, (2)~filter for predictive consistency, and (3)~distill into a student model. Readers primarily interested in the practical pipeline may consult Fig.~\ref{fig_PARROT} and return to the derivation for justification.
\begin{tcolorbox}[
  enhanced,
  boxrule=0pt,
  frame hidden,
  borderline west={3pt}{0pt}{red!70!black},
  colback=red!5,
  sharp corners,
  left=10pt, right=10pt, top=6pt, bottom=6pt
]
\textbf{Takeaway 1: Why Reasoning Rewards Resist Reward Hacking.}
Scalar reward models are susceptible to reward hacking because they compress evaluation into a single number that can be inflated by exploiting biases without genuine quality improvement. As shown in Fig.
\ref{fig_reward_hacking_example}, RL with scalar rewards exhibits reward increases while generation quality visibly degrades. \model mitigates this through an implicit regularization: the model must produce coherent, multi-dimensional reasoning \textit{before} emitting scores, structurally grounding evaluation in interpretable criteria. Reward inflation without corresponding visual evidence becomes difficult when the model must justify \emph{why} each dimension merits its score. Empirically, \model maintains monotonic correspondence between reward and quality throughout training (Fig.\ref{fig_reward_hacking_example}, Fig.\ref{fig:rl_curve}). We envision that \model will facilitate broader and more reliable research on diffusion RL methods.
\end{tcolorbox}

\subsection{Variational Framework: The Hindsight-Foresight Decomposition}
\label{sec:PARROT}
Let $x = (I_A, I_B, c)$ denote a comparison tuple comprising two generated images and a conditioning user request $c$ (which includes text instructions and, for editing tasks, a source image). Let $y \in \{A \succ B, B \succ A\}$ denote the ground-truth human preference. Unlike reward models that model $P(y|x)$ directly, PARROT introduces a latent natural language rationale $z$ that \textit{explains} preference $y$.

We treat the rationale $z$ as the explanatory mechanism underlying the preference. We use ``explanatory'' in the sense of predictive sufficiency: $z$ is a valid rationale if it contains sufficient information to predict preference $y$ from evaluation task $x$. Our goal is to learn a evaluator reward model (the \textit{Student}) $P_{\theta}(z, y | x)$ capable of generating the rationale $z$ and predicting the preference $y$ for downstream tasks. To learn this from preference data alone, we maximize the Evidence Lower Bound (ELBO):

\begin{equation}
\mathcal{L}_{\text{ELBO}} = 
    \underbrace{\mathbb{E}_{z \sim q_{\phi}} [\log P_{\theta}(y|x, z)]}_{\text{Term 1: Prediction}}  - \underbrace{D_{\text{KL}}\bigl(q_{\phi}(z|x, y) \,\|\, P_{\theta}(z|x)\bigr)}_{\text{Term 2: Regularization}} 
    \label{eq_distill}
\end{equation}

\methodfigure
\rmqualityfigure
This derivation reveals a natural ``Teacher-Student'' structure, decomposing the learning process into two complementary modes:
\begin{itemize}[leftmargin=15pt, itemsep=0pt, topsep=0pt]
    \item \textbf{Hindsight (Posterior $q_{\phi}(z|x,y)$):} Inferring the rationale $z$ when the ground-truth preference $y$ is \textit{known}---analogous to how human experts articulate evidence after forming an initial judgment.
    \item \textbf{Foresight (Prior $P_{\theta}(z, y|x)$):} Predicting both rationale $z$ and preference $y$ from the input $x$ alone---our target rationalized reward model.
\end{itemize}

\textbf{Phase 1: Rationale Generation} (Constructing $q_{\phi}(z|x,y)$). \quad
A naive approach prompts a teacher VLM to compare images without guidance, sampling from the prior $p(z|x)$. This is suboptimal: even strong VLMs frequently misjudge subtle visual details (e.g., Table~\ref{tab:main_results} shows even Gemini-3-Pro has 30\% disagreement with human preferences). Instead, we use \textbf{preference anchoring}: the Teacher (Qwen3-VL-32B-Instruct) generates rationales \textit{conditioned on} the known preference label $y$, collapsing generation from open-ended evaluation to focused justification. This concentrates probability mass on rationales consistent with the observed label, yielding higher-quality posterior samples than unconditioned generation---confirmed empirically in Table~\ref{tab:main_results}. Brief prompt templates are shown below.
\begin{tcolorbox}[promptbox, title={\color{white}Rationale Generation (Phase 1)}]
\textbf{Input:} Instruction \texttt{\{inst\}}, Source Image, Edited Image A, Edited Image B\\[1pt]
\textbf{Task:} Compare two edited images according to the instruction.\\[2pt]
\textbf{Aspects} (each scored 1--4): \textit{Text Faithfulness} $\cdot$ \textit{Image Faithfulness} $\cdot$ \textit{Physical \& Visual Quality} $\cdot$ \textit{Text Rendering}\\[2pt]
\textbf{Preference Anchor:} ``Hint: human preference is: \texttt{\{label\}}''\\[2pt]
\textbf{Output Format:}\\[-2pt]
{\footnotesize
\texttt{[Understanding of the user request]}\\
\texttt{\# Detailed Judgement}\\
\texttt{1. Text Faithfulness:}\\
\texttt{~~\#\# Justification: [...] \#\# Score A: [...] \#\# Score B: [...] \#\# Winner: [...]}\\
\texttt{2--4. \textit{(same structure for remaining aspects)}}\\
\texttt{\# Summary: [...]}
}
\end{tcolorbox}

\textbf{Phase 2. Predictive Consistency Filtering: Maximizing Term~1, $\mathbb{E}_{z \sim q_{\phi}}[\log P(y|x,z)]$.}
 
While Phase~1 produces rationales that are linguistically plausible, plausibility does not guarantee predictive sufficiency. A rationale $z$ contributes to the ELBO only if it successfully explains $y$; otherwise, $\log P(y|x,z)$ is low and the corresponding sample degrades the bound. For instance, a VLM might generate a rationale that sounds correct in isolation (e.g., ``Image B has distorted text'') but does not align with the visual content, or it may ignore the provided preference label altogether.

\qualityfigure
To enforce predictive sufficiency and thereby maximize Term~1, We enforce that rationales actually explain the preference via a consensus check: the Teacher is re-queried with $z$ \textit{without} the preference label, verifying that $z$ alone suffices to recover $y$:
\begin{equation}
\vspace{-.1cm}
    \mathcal{C}(x, y, z) = \mathbb{I} \left[ \operatorname*{arg\,max}_{y'} P_{\text{Teacher}}(y' | x, z) = y \right]
\end{equation}
We retain $(x, y, z)$ only if $\mathcal{C}=1$, yielding filtered dataset $\mathcal{D}_{\text{pair}}$. This approximates maximizing $\mathbb{E}_{q}[\log P(y|x,z)]$ by restricting $q_\phi$'s support to the high-likelihood region, discarding hallucinated or insufficiently informative rationales.

\textbf{Phase 3. Foresight Learning: Minimizing Term~2 $D_{\text{KL}}(q_{\phi}(z|x, y) \,\|\, P_{\theta}(z|x))$}
\label{sec:pointwise_projection}
We train the Student $P_{\theta}(z|x)$ to generate rationales \textit{without} the preference label via SFT on filtered posterior samples. Since $q_\phi$ is fixed, minimizing the KL reduces to maximizing $\mathbb{E}_{q_{\phi}}[\log P_{\theta}(z|x)]$---precisely the standard SFT objective on filtered samples.

\textbf{Bridging Pairwise Training and Pointwise Deployment.} While we derive the ELBO from pairwise data (which is easier to collect), downstream applications require \textit{pointwise} feedback, e.g., scalar scores for RL training, critiques on individual images for test-time prompt refinement, visual grounding for diagnostic and dense visual rewards.  A model trained solely on pairwise comparisons often fails to critique a single image in isolation, as it overfits to the presence of a contrastive candidate.

We address this with a \textbf{Pointwise Projection Strategy}, based on the assumption that pairwise and pointwise assessment share common evaluation principles. We prompt the Teacher to assess each image in isolation, providing the validated pairwise rationale $z_{\text{pair}}$ as a reference hint to guide attention toward identified defects. The Teacher articulates absolute scores on a 1--4 scale (with float granularity) across four dimensions: Text Faithfulness, Image Faithfulness, Physical Quality, and Text Rendering. Detailed rubrics are in the appendix. This projection extends beyond the strict pairwise ELBO, but the projected rationales inherit their quality from the ELBO-filtered pairwise rationales and maintain the same predictive relationship between reasoning and scores.

This induces a pointwise dataset $\mathcal{D}_{\text{point}}$. We train the Student jointly on both datasets to enable both pointwise and pairwise assessments: 
\( \mathcal{L}_{\text{SFT}} 
    = \mathbb{E}_{(x,y,z)\sim \mathcal{D}_{\text{point}}\cup\mathcal{D}_{\text{pair}}}
    \Bigl[ -\log P_{\theta}(z , y  | x ) \Bigr] \)

\begin{tcolorbox}[
  enhanced,
  boxrule=0pt,
  frame hidden,
  borderline west={3pt}{0pt}{red!70!black},
  colback=red!5,
  sharp corners,
  left=10pt, right=10pt, top=6pt, bottom=6pt
]
\textbf{Takeaway 2: Trained Reward Models vs.\ Generic VLM Judges.}
A natural question is why not directly use a capable generic VLM (e.g., Qwen3-VL-32B) as a judge. Beyond the practical advantage of a smaller 8B model, we identify a more fundamental reason: \emph{scoring stability}. Generic VLMs produce high-variance scores when used as pointwise evaluators---two images of comparable quality may receive markedly different scores across independent queries, introducing noise that destabilizes RL optimization. This variance arises because generic VLMs are not calibrated for fine-grained quality discrimination. In contrast, preference training via \framework explicitly calibrates the model to produce \emph{low-variance, preference-aligned scores} (Fig.~\ref{fig:rl_curve}): it learns to reason in a way that is predictive of human preference, yielding stable rewards across semantically equivalent inputs. Tables \ref{tab:unigen_ablation} and \ref{tab:ablation_edit} confirm this empirically: despite being 4$\times$ smaller, \model consistently outperforms the generic Qwen3-VL-32B judge as an RL reward signal across all tested generators.
\end{tcolorbox}

\subsection{From Evaluator to Optimizer: Tuning in Parameter Space and Prompt Space}

The rationalized reward model enables optimization in two complementary spaces, each suited to various deployment scenarios.

\begin{itemize}[leftmargin=*,itemsep=0pt, topsep=0pt, parsep=2pt, partopsep=1pt]
    \item \textbf{Parameter Space (SFT/RL Fine-Tuning).} Multi-dimensional scores provide semantically decomposed reward signals for reinforcement learning, enabling fine-grained feedback across quality dimensions rather than optimization against a single opaque scalar. The structured rationales further serve as natural-language explanations for reward assignments, aiding interpretability and reducing reward hacking (see Fig.~\ref{fig_reward_hacking_example}). 
    
    \item \textbf{Prompt Space (Test-Time Refinement).} Natural-language rationales identify concrete deficiencies in generated images, which we leverage to construct a \textbf{Generate--Critique--Refine} loop (Fig.~\ref{fig_inference}): \model critiques an initial generation, and its critique is used to produce a targeted prompt revision for re-generation. This performs $t^* = \arg\max_t R(G(t))$ guided by language rather than numerical gradients, trading test-time compute for quality without parameter updates~\citep{snell2024scaling}. We note that this post-hoc prompt refinement dataset also enables distillation for pre-hoc prompt enhancement models. 
\end{itemize}

\noindent This dual-space formulation connects to test-time compute scaling~\citep{snell2024scaling}: prompt-space optimization offers an axis for improving generation quality orthogonal to parameter-space training and applicable to any frozen generator. We hypothesize that it is particularly effective when the generator possesses latent capabilities under-elicited by suboptimal prompts---a working hypothesis we examine empirically in Section~\ref{sec:experiments}.
 \inferencefig
\begin{tcolorbox}[
  enhanced,
  boxrule=0pt,
  frame hidden,
  borderline west={3pt}{0pt}{red!70!black},
  colback=red!5,
  sharp corners,
  left=10pt, right=10pt, top=6pt, bottom=6pt
]
\textbf{Takeaway 3: Why Reasoning Rewards Enable Effective Test-Time Scaling.}
Generators often possess latent capacity for high-quality outputs that is under-elicited by suboptimal prompts. \model unlocks this capacity without weight modification---but why is a preference-trained reward model more effective here than a generic prompt rewriter~\citep{promptenhancer}? The key difference is that \model has internalized an objective aligned with human judgment: when it critiques an image and proposes a revision, it optimizes in prompt space toward \emph{maximizing human preference likelihood}, targeting precisely the dimensions where the output deviates from what humans would prefer. Unlike blind prompt enhancement that rewrites without observing outputs, \model performs post-hoc, preference-aware refinement---observing the actual failure and producing targeted corrections. This explains why a single Generate--Critique--Refine iteration can match or surpass RL fine-tuning for generators with strong latent capabilities (Section~
\ref{sec:experiments}).
\end{tcolorbox}

\section{Experiments}
\label{sec:experiments}

\paragraph{Training Data.}
We evaluate \model on both image generation and image editing tasks. Our training data derives from existing preference datasets: 30K query-preference pairs from EditReward~\citep{editreward} for image editing, and 50K pairs from HPDv3 and RapidData~\citep{hpdv3} for text-to-image generation. These datasets provide only binary or ranked preference labels without explanations. We apply the \framework pipeline (\S
~\ref{sec:PARROT}) with Qwen3-VL-32B-Instruct as the teacher model to transform these raw preference pairs into reasoning-annotated training data. Our data scale is 10-20 times smaller -- this efficiency stems from the teacher model's pre-trained knowledge, which PARROT distills through structured rationales rather than raw labels; the ablation in \S
~\ref{sec:ablation} isolates this factor. During Phase~2 (consistency filtering), approximately 72\% of generated rationales survive the predictive consistency check, indicating that preference anchoring produces largely coherent rationales while the filter removes a meaningful fraction of hallucinated or insufficiently informative samples. Full implementation details (training hyperparameters, hardware configuration, RL setup) are provided in Appendix. \textbf{All code, data, and models are released at \faGlobe\ 
\href{https://tiger-ai-lab.github.io/RationalRewards/}{{\textbf{Project Page}}} to facilitate reproducibility and further research.}
 
\subsection{Accuracy in Preference Modeling}\label{sec:ablation}

\maintable

We first evaluate whether \model produces human-aligned preference judgments. We report pairwise comparison accuracy on three established benchmarks: Multimodal Reward Bench 2~\citep{mmrb2} and GenAI-Bench~\citep{jiang2024genai}  and EditReward Bench~\citep{editreward} for both text- and image-to-image generation.

\textbf{Main Results.}
As shown in Table~\ref{tab:main_results}, our 8B-parameter \model surpasses all open-source scalar reward models by a substantial margin across all three benchmarks, without requiring complex loss designs to handle label noise or annotation ambiguities. Notably, \model outperforms commercial models including Gemini-2.5-Flash and approaches the performance of GPT-5/Gemini-2.5-Pro on preference prediction, offering a cost-effective alternative for quality assessment and evaluation in visual generation.
 
\textbf{Ablation of PARROT versus Direct Distillation.}
To isolate the contribution of \framework from generic knowledge distillation, we include a baseline that performs direct SFT distillation from Qwen3-VL-32B-Instruct to the same 8B backbone, using the same data volume but without preference-anchored rationalization (marked ``Qwen3-VL-32B-Instruct Distillation'' in Table~
\ref{tab:main_results}). This baseline underperforms \model on all benchmarks---by 6.8 points on MMRB2 (T2I) and 17.3 points on GenAI (Edit)---confirming that the structured rationalization process, not simply access to a larger teacher, drives the performance gains. We also replace the backbone with Qwen2.5-VL-7B-Instruct; the results still exceed prior scalar reward models, clarifying that improvements are attributable to \framework rather than the specific choice of backbone.

\subsection{Optimization in Dual Spaces}
\label{sec:dual_space_exp}

Given the strong discriminative performance of \model, we now investigate its utility for improving downstream generation. We explore two complementary optimization strategies: \textit{parameter-space} tuning via RL and \textit{prompt-space} tuning via test-time critique-and-refinement. We evaluate on ImgEdit-Bench~\citep{imgedit} and GEdit-Bench-EN~\citep{step1edit} for image editing, the UniGen benchmark for text-to-image generation. We also include in the appendix a physics-centric PICA-Bench~\citep{picabench} for out-of-distribution stress testing, following each benchmark's prescribed evaluation protocol.


\textbf{Parameter Space Tuning (RL).}
We experiment with the recent Diffusion RL approach, DiffusionNFT~\citep{diffusionnft}, which samples a group of generations for the same user prompt and optimizes with a weighted diffusion loss. For reproducibility, we include the algorithm and implementation details in the appendix. 
We use \model to provide dense, per-dimension reward signals for RL fine-tuning and systematically compare against alternative reward models spanning two axes: \emph{scalar vs.\ reasoning-based} and \emph{generic vs.\ preference-trained}:
\begin{enumerate}[nosep,leftmargin=*,itemsep=0pt, topsep=0pt, parsep=2pt, partopsep=1pt]
    \item \textbf{Scalar reward models}: EditReward~\citep{editreward} for image editing and MultiReward (used by DiffusionNFT~\citep{diffusionnft}) for text-to-image generation. These output a single scalar score without natural language reasoning.
    \item \textbf{Generic reasoning model}: Qwen3-VL-32B-Instruct used directly as a judge. This model can produce natural language critiques but has \emph{not} been trained on preference data via \framework, isolating the contribution of our training pipeline from raw model scale.
\end{enumerate}
\unigenablationtable

As shown in Tables~\ref{tab:ablation_sidebyside} and~\ref
{tab:unigen_ablation}, RL with \model yields consistent improvements over both base models across nearly all subcategories, surpassing both scalar reward baselines and the generic reasoning baseline. For image editing, \model-guided RL improves Flux.1 Kontext from 3.52 to 3.84 overall on ImgEdit-Bench, outperforming EditReward-guided RL (3.66) by a clear margin. For text-to-image generation, \model lifts FLUX.1-dev from 60.97 to 70.34 on UniGen (+9.37 points), substantially exceeding both MultiReward (62.55) and the direct Qwen3-VL-32B judge (66.71). Notably, the 8B \model outperforms Qwen3-VL-32B used as a direct judge, confirming that PARROT's structured preference training provides value beyond raw model capacity. 
 
\editablationtable

\textbf{Test-Time Prompt Space Tuning.}
 We leverage the generative nature of \model in a \textit{Generate--Critique--Refine} protocol: the generator produces an initial image; \model evaluates it across four dimensions with natural language critique and refinement suggestions; if any dimension score falls below a threshold of 3.0, the refined request is fed back to the generator. This single-iteration loop adds approximately 0.4 seconds of VLM inference overhead per image (via vLLM prefix caching and paged attention), compared to ${\sim}$384 GPU-hours for RL fine-tuning of a single base model.


\textbf{Prompt Tuning Matches or Exceeds RL.}
A striking finding emerges from Table~
\ref{tab:ablation_sidebyside}: inference-time prompt tuning frequently yields improvements comparable to or exceeding computationally expensive RL. On ImgEdit-Bench, prompt tuning boosts the RL-tuned Flux model from 3.84 to 4.01 overall. For Qwen-Image-Edit, prompt tuning applied on top of RL yields the best overall score of 4.43, with the two methods proving complementary. On GEdit-Bench-EN Overall, prompt tuning (8.33) slightly exceeds RL alone (8.29).

The RL performance ceiling is partly structural: LoRA-based fine-tuning constrains parameter update capacity, and the RL query distribution may not fully cover the evaluation distribution. In contrast, prompt tuning performs per-instance optimization without risk of catastrophic forgetting. More fundamentally, these results suggest a \emph{latent capability hypothesis}: generators already possess the capacity for high-quality outputs, but this capacity is under-elicited by suboptimal prompts. \model's critique bridges user intent and model capability without weight modification. We note this remains a hypothesis requiring representation-level validation.
 
\vspace{-.2cm}
\section{Related Work}
\textbf{Reward Models for Visual Generation.} The standard paradigm in visual generation relies heavily on scalar reward models trained on large-scale human preference datasets. Models such as ImageReward~\citep{imagereward},VideoReward~\citep{liu2025improving}, PickScore~\citep{kirstain2023pick}, UnifiedReward~\citep{unifiedreward} and EditReward~\citep{editreward} typically function as opaque discriminators, mapping pixel inputs directly to a scalar score. Our work provides an alternative path for reward modeling, shifting the paradigm from scalar regression to rationalization~\citep{star}. Generative reward models have also been studied in verifiable domains~\citep{genrm, rrm, chen2026rmr1rewardmodelingreasoning}.

\textbf{Training and Test-Time Scaling in Visual Generation.} Recent efforts, such as FlowGRPO~\citep{flowgrpo}, DanceGRPO~\citep{xue2025dancegrpo}, Blip3o-Next~\citep{chen2025blip3o}, and DiffusionNFT~\citep{diffusionnft, editr1}, successfully integrated RL into visual generation, demonstrating significant gains in compositional reasoning and text rendering. While effective, RL is bottlenecked by the quality of the reward model, often suffering from reward hacking when the proxy reward diverges from human preference. Recent works have pivoted toward trading test-time compute for enhanced generation quality. ReflectionFLow~\citep{zhuo2025reflection} and PromptEnhancer~\citep{promptenhancer} utilizes a Chain-of-Thought (CoT) rewriter to expand user prompts into detailed specifications prior to generation. For image editing, Reason-Edit~\citep{reasonedit} introduces a thinking--editing--reflection loop. Most recently, several approaches have begun leveraging the multimodal CoT capabilities of Unified Multimodal Models to iteratively improve visual synthesis at test time~\citep{qin2025uni, wu2025omnigen2, deng2025emerging, jiang2025t2i, ye2025visual, editr1}. Our work highlights the importance of preference calibration and rationalization in reward models, revealing the fundamental mechanism of trading test-time compute for better generation. 

\vspace{-.2cm}
\section{Conclusions}
\usagefiguretwo

We presented \model, a reasoning-based reward model that replaces opaque scalar scoring with structured, multi-dimensional chain-of-thought critiques, and \framework, a variational framework that makes this tractable by treating rationales as latent variables recoverable from readily available preference data. Our work yields three principal findings. First, structured rationalization acts as a powerful inductive bias: by requiring the model to articulate why one image is preferred, an 8B-parameter model achieves preference-prediction accuracy competitive with Gemini-2.5-Pro and approaching GPT-5, while consuming 10–20× less training data than scalar baselines. Second, the multi-dimensional rationales produced by \model serve as semantically grounded RL rewards that consistently outperform both scalar reward models and generic VLM judges of larger scale across text-to-image and image-editing benchmarks. Third, and most notably, the Generate–Critique–Refine loop -- a purely test-time intervention requiring no parameter updates -- matches or exceeds RL-based fine-tuning on several benchmarks, lending empirical support to the hypothesis that current generators harbor latent capabilities that suboptimal prompts fail to elicit.

\bibliography{colm2026_conference}
\bibliographystyle{colm2026_conference}

\input{appendix.tex}

\end{document}

%% file: math_commands.tex
\usepackage{amsmath,amsfonts,bm}

\def\eqref#1{equation~\ref{#1}}

\def\1{\bm{1}}

\DeclareMathAlphabet{\mathsfit}{\encodingdefault}{\sfdefault}{m}{sl}
\SetMathAlphabet{\mathsfit}{bold}{\encodingdefault}{\sfdefault}{bx}{n}



%% file: figures_def.tex

\def\usagefigure{
\begin{figure*}[t]
    \centering
    \includegraphics[width=0.85\textwidth]{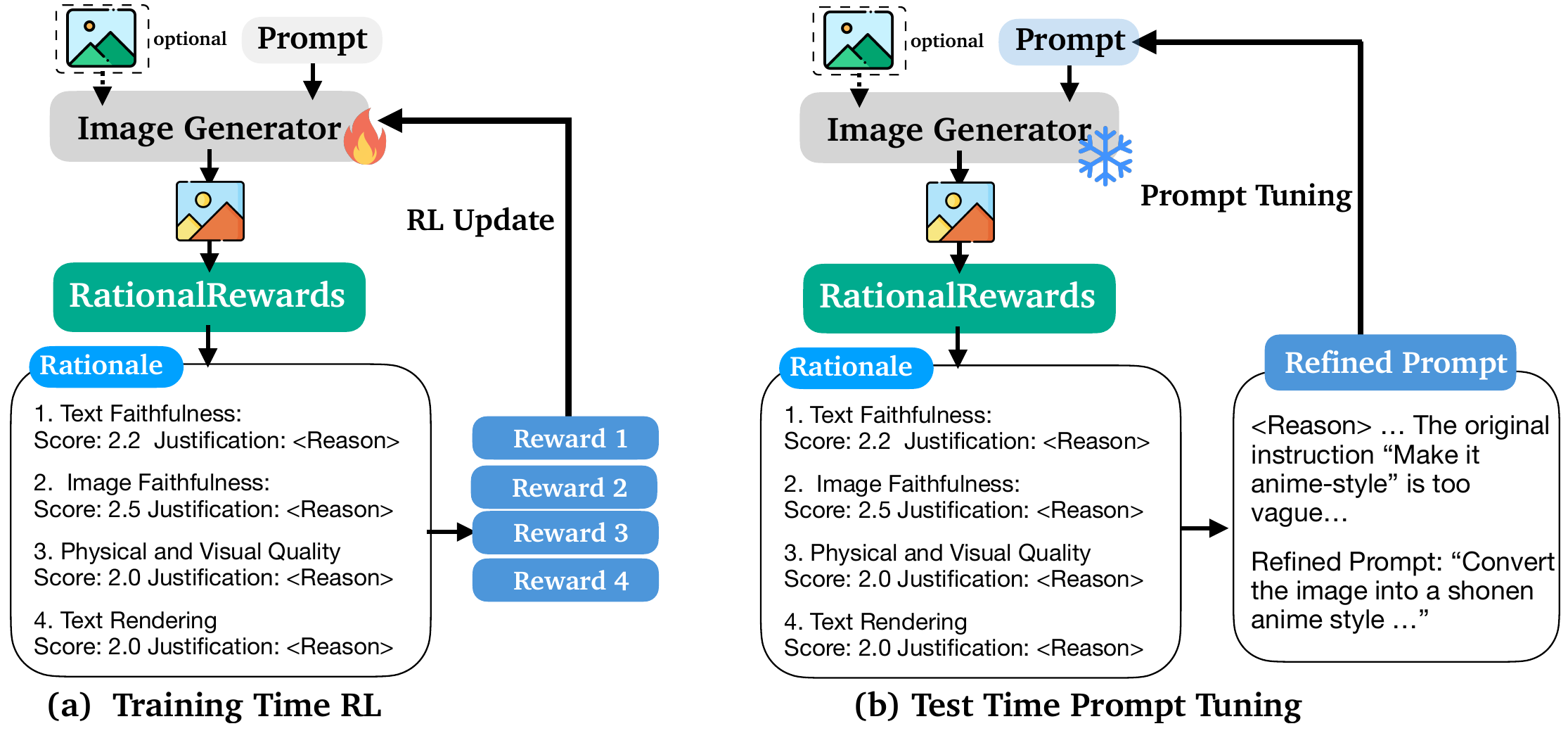}
    \caption{\small \model is a reasoning-based reward model that produces structured rationales before assigning scores, enabling dual-space optimization for image generation.
    (a) As a reward model, it improves RL-based fine-tuning of generators over scalar baselines;
    (b) as a test-time optimizer, its Generate--Critique--Refine loop matches or surpasses RL-based optimization on multiple benchmarks without parameter updates.}
    \label{fig:teaser}
\end{figure*}
}

\def\usagefiguretwo{
\begin{figure*}[t]
    \centering
    \includegraphics[width=0.85\textwidth]{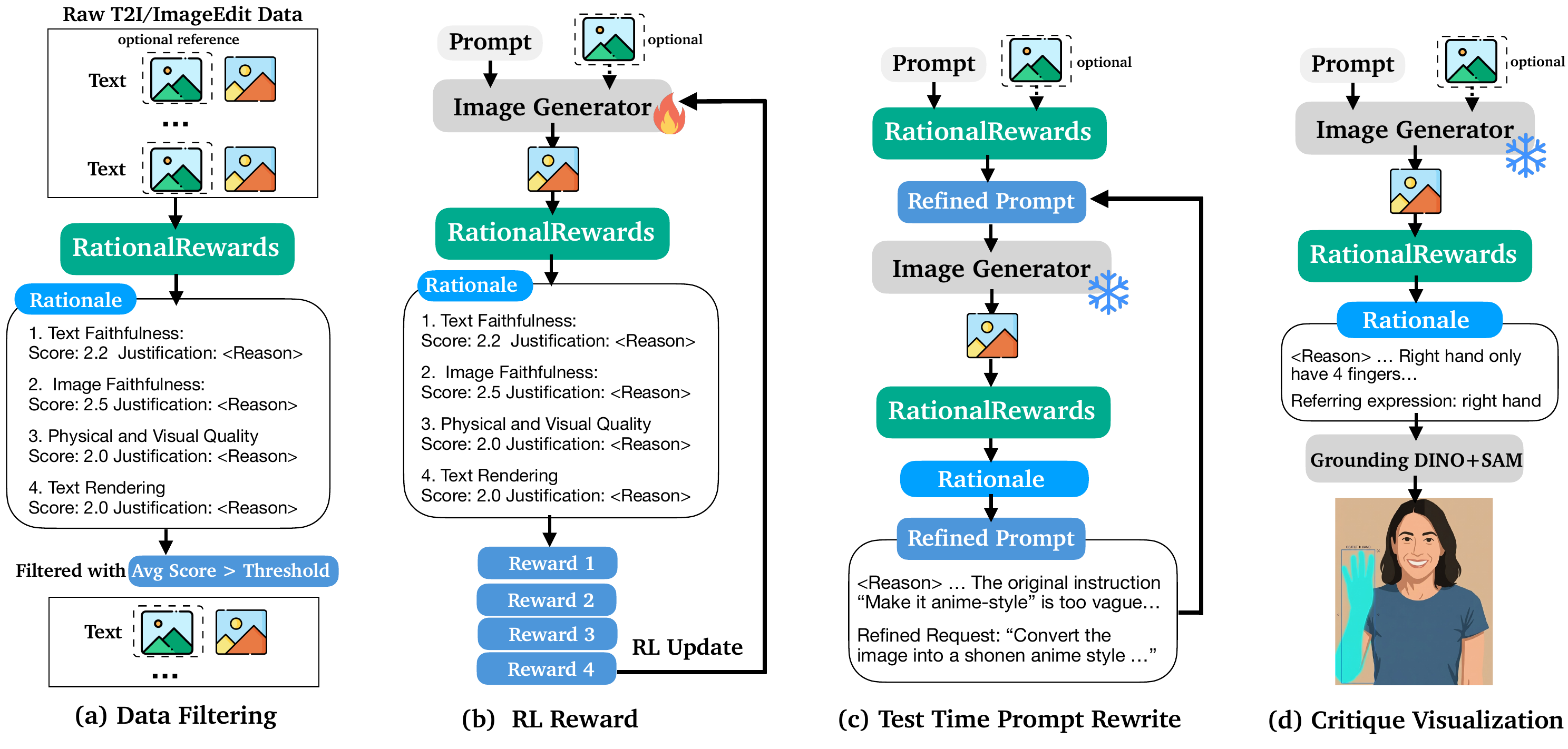}
    \caption{\small \model  (a) enables explainable quality control for data curation; 
    (b) serves as a multi-dimensional reward model driven by transparent rationales;
    (c) serves as a preference-calibrated test-time prompt tuner that trades compute for better generation quality; (d) fuels regional flaw grounding and dense visual rewards. }
    \label{fig_usage}
\end{figure*}
}

\def\rewardhackfigure{\begin{figure*}[t]
    \centering
    \includegraphics[width=\textwidth]{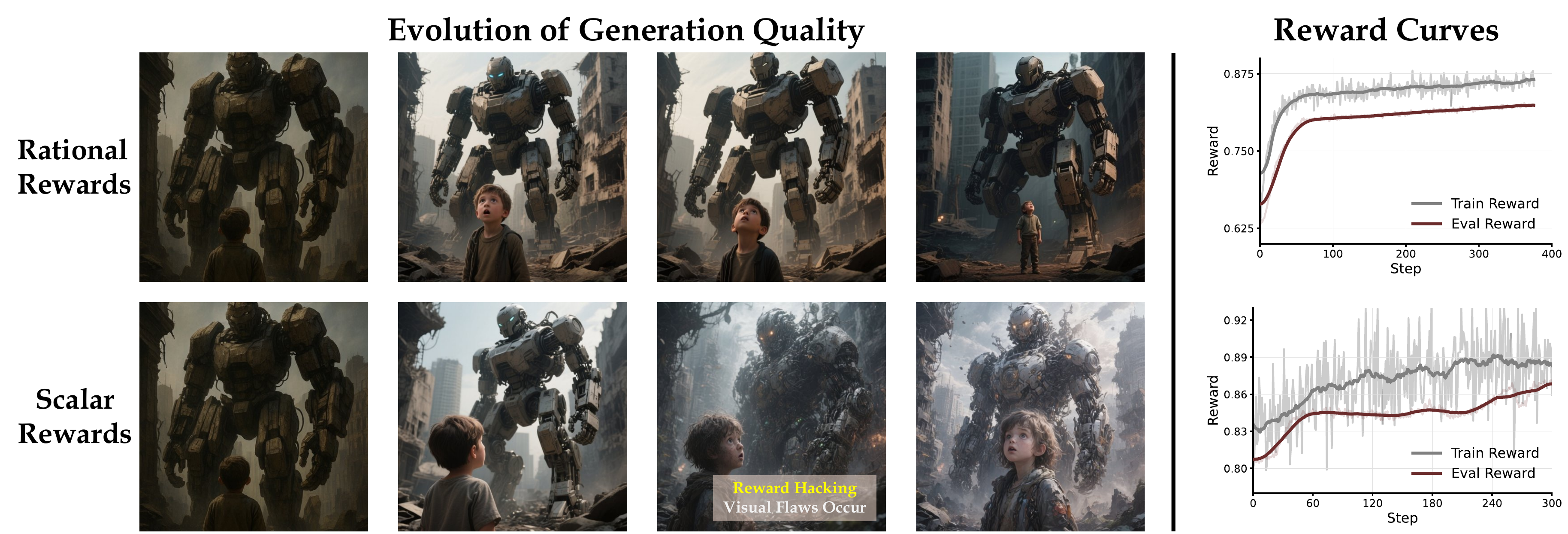}
    \caption{\small RL (LoRA) training on Qwen-Image using scalar rewards encounter reward hacking (bottom row): as training reward continues to grow, generation quality starts to degenerate, because black box rewards mislead visual generators with biases. In contrast, \model (top row) sustains generation quality with stable reward growth. See Fig.~10 and 11 for more details.}
    \label{fig_reward_hacking_example}
\end{figure*}}

\def\ablationdata{
\midrule
Flux.1 Kontext [dev] & 3.52 & 7.16 & 7.37 & 6.51 \\
+RL (EditReward) & 3.66 & 7.38 & 7.53 & 6.88 \\

\rowcolor{yellow!20} +RL (\model) & 3.84 & 7.75 & 8.24 & \textbf{7.37} \\
+PT (\model) & \textbf{4.01} & 7.77 & 7.61 & 7.23 \\
\midrule
Qwen-Image-Edit & 4.27 & 8.00 & 7.86 & 7.56 \\
+RL (EditReward) & 4.25 & 8.36 & 7.91 & 7.77 \\
+RL (Qwen3-VL-32B) & 4.25 & 8.42 & 7.83 & 7.79 \\

\rowcolor{yellow!20} +RL (\model) & 4.38 & 8.74 & 8.43 & 8.29 \\
+PT (\model) (Test Time) & \textbf{4.43} & 8.94 & 8.20 & \textbf{8.33} \\
\bottomrule
}

\def\ablationheader{
\toprule
\multirow{2}{*}{\textbf{Model}} & \textbf{ImgEdit} & \multicolumn{3}{c}{\textbf{GEdit-Bench-EN}} \\
\cmidrule(lr){2-2} \cmidrule(lr){3-5}
 & \textbf{Overall} & \textbf{G\_SC} & \textbf{G\_PQ} & \textbf{G\_O} \\
}

\def\ablationcaption{Ablation of \model on downstream image editing benchmarks.}

\def\ablationwrapfig{
\begin{wraptable}{r}{0.42\textwidth}
\vspace{-0.4cm}
\centering
\caption{\small \ablationcaption}
\label{tab:ablation_wrap}
\resizebox{0.42\textwidth}{!}{%
\begin{tabular}{lcccc}
\ablationheader
\ablationdata
\end{tabular}%
}
\vspace{-0.4cm}
\end{wraptable}
}

\def\ablationtable{
\begin{table}[t!]
\centering
\caption{\small \ablationcaption}
\label{tab:ablation}
\resizebox{\columnwidth}{!}{%
\begin{tabular}{lcccc}
\ablationheader
\ablationdata
\end{tabular}%
}
\end{table}
}

\def\unigenablationtable{
\begin{table*}[t!]
\centering
\caption{\small Ablation of \model for Text-to-image RL on UniGenBench++. We compare scalar reward model~\textit{MultiReward} and generic reasoning reward~ \textit{Qwen3-VL-32B}.}

\resizebox{\textwidth}{!}{%
\begin{tabular}{lcccccccccc|c}
\toprule
\textbf{Model} & \textbf{Action} & \textbf{Attribute} & \textbf{Compound} & \textbf{Layout} & \textbf{Grammar} & \textbf{Logic} & \textbf{Relation} & \textbf{Style} & \textbf{Text} & \textbf{World Know.} & \textbf{Overall} \\
\midrule
FLUX.1-dev               & 62.24 & 67.20 & 45.75 & 70.84 & 62.30 & 29.77 & 66.88 & 85.00 & 32.18 & 87.50 & 60.97 \\
+MultiReward             & 59.78 & 68.23 & 44.21 & 74.37 & 59.33 & 28.25 & 68.35 & 76.05 & 36.21 & 86.03 & 60.12 \\
+Qwen3-VL-32B            & 65.47 & 72.68 & 53.28 & 71.82 & 60.78 & 33.24 & 71.85 & 85.53 & 42.15 & 89.47 & 66.53 \\
\rowcolor{yellow!20} +RationalRewards         & 67.40 & 76.36 & 57.67 & 72.15 & 60.29 & 40.53 & 74.59 & 87.20 & 52.57 & 90.61 & \textbf{70.34} \\
\midrule
SD-3.5-Medium            & 60.41 & 66.99 & 53.35 & 70.31 & 59.89 & 37.73 & 68.78 & 89.80 & 15.23 & 84.34 & 60.71 \\
\rowcolor{yellow!20} +RationalRewards         & 64.36 & 81.49 & 67.98 & 75.88 & 58.68 & 42.37 & 75.60 & 89.60 & 10.05 & 91.77 & \textbf{70.56} \\
+MultiReward             & 57.03 & 66.67 & 51.03 & 75.37 & 57.22 & 34.86 & 67.51 & 77.60 & 21.84 & 86.71 & 62.55 \\
+Qwen3-VL-32B            & 61.23 & 74.48 & 63.85 & 75.34 & 59.67 & 31.23 & 72.84 & 84.73 & 14.87 & 88.86 & 66.71 \\
\midrule
Qwen-Image               & 82.49 & 87.93 & 72.94 & 86.56 & 60.96 & 51.59 & 80.08 & 94.70 & 72.13 & 94.15 & 78.36 \\
+MultiReward             & 79.52 & 86.45 & 70.91 & 88.53 & 58.43 & 48.62 & 80.55 & 83.75 & 67.18 & 92.17 & 75.61 \\
+Qwen3-VL-32B            & 81.95 & 87.45 & 76.42 & 87.73 & 62.93 & 51.14 & 81.55 & 95.20 & 75.67 & 95.63 & 80.17 \\
\rowcolor{yellow!20} +RationalRewards         & 82.11 & 87.82 & 78.82 & 88.07 & 66.21 & 52.88 & 82.21 & 96.60 & 79.76 & 96.57 & \textbf{82.60} \\
\bottomrule
\end{tabular}%
}
\vspace{-.3cm}
\label{tab:unigen_ablation}
\end{table*}
}

\def\maintable{
\begin{table*}[t!]
\centering
\caption{Comparison of reward models as evaluators. We include Multimodal Reward Bench~2 (MMRB2), EditReward-Bench, and GenAI-Bench. T2I and Edit means text-to-image and image-to-image respectively.}
\label{tab:main_results}
\resizebox{\textwidth}{!}{%
    \begin{tabular}{lccccc}
    \toprule
    \multirow{2}{*}{\textbf{Judge}} 
        & \multicolumn{2}{c}{\textbf{MMRB2}} 
        & \multirow{2}{*}{\textbf{EditReward}} 
        & \multicolumn{2}{c}{\textbf{GenAI-Bench}} \\
    \cmidrule(lr){2-3} \cmidrule(lr){5-6}
        & \textbf{T2I} & \textbf{Edit} 
        & 
        & \textbf{T2I} & \textbf{Edit} \\
    \midrule
    Qwen2.5-VL-7B~\citep{bai2025qwen2} & 50.4 & 57.1 & 52.69 & -- & 40.48 \\
    Qwen2.5-VL-72B & 59.1 & 64.6 & 63.9 & 66.6 & 74.3 \\
    Qwen3-VL-8B~\citep{qwen3} & 59.4 & 61.7 & 51.9 & 55.1 & 50.1 \\
    Qwen3-VL-32B & 64.1 & 67.3 & 64.2 & 66.9 & 76.3 \\
    \midrule
    EditReward-7B~\citep{editreward} & -- & 67.2 & 56.99 & -- & 65.72 \\
    UnifiedReward-7B~\citep{unifiedreward} & 59.8 & -- & -- & 67.9 & -- \\
    \textbf{\model (Qwen2.5-VL-7B)} & 62.3 & 68.5 & 63.6 & 66.4 & 75.7 \\
    \rowcolor{yellow!20} \textbf{\model (Qwen3-VL-8B)} & \textbf{64.2} & \textbf{70.3} & \textbf{66.2} & \textbf{69.8} & \textbf{80.1} \\
    Qwen3-VL-32B-Instruct Distillation & 57.4 & 65.6 & 56.8 & 59.3 & 62.8 \\
    \midrule
    \multicolumn{6}{l}{\textit{\textbf{Commercial Models}}} \\
    \midrule
    GPT-4.1 & 65.8 & 68.2 & 58.3 & 60.5 & 69.3 \\
    Gemini 2.5 Flash~\citep{gemini25} & 63.1 & 66.5 & 58.6 & 65.8 & 73.0 \\
    Gemini 2.5 Pro & 70.5 & 71.3 & 71.3 & 66.2 & 78.9 \\
    Gemini 3 Pro~\citep{gemini3} & 74.4 & 74.9 & 72.2 & 73.1 & 80.5 \\
    \bottomrule
    \end{tabular}%
}
\end{table*}
}

\def\unigentable{
\begin{table*}[t!]
\centering
\caption{\small Text-to-image generation results on UniGen benchmark. We report category-level scores and overall performance. \textbf{Action} is the average of Hand, Full Body, Animal, Non Contact, Contact, and State. \textbf{Layout} is the average of 2D and 3D.}
\resizebox{\textwidth}{!}{%
\begin{tabular}{lcccccccccc|c}
\toprule
\textbf{Model} & \textbf{Action} & \textbf{Attribute} & \textbf{Compound} & \textbf{Layout} & \textbf{Grammar} & \textbf{Logic} & \textbf{Relation} & \textbf{Style} & \textbf{Text} & \textbf{World Know.} & \textbf{Overall} \\
\midrule
\multicolumn{12}{c}{\textit{Open-source / Commercial Models}} \\
\midrule
Nano Banana Pro~\citep{nanobanana}          & 91.30 & 91.95 & 92.91 & 93.30 & 89.59 & 80.24 & 95.43 & 99.30 & 95.65 & 97.47 & 92.72 \\
Seedream-4-5-251128      & 88.06 & 91.03 & 90.08 & 92.55 & 84.09 & 73.17 & 90.61 & 99.20 & 91.67 & 96.35 & 89.70 \\
UniWorld-V1~\citep{uniworld}              & 66.94 & 70.62 & 54.51 & 68.96 & 63.77 & 38.41 & 67.13 & 91.10 & 26.44 & 82.91 & 63.11 \\
OmniGen2~\citep{omnigen}                 & 62.68 & 72.12 & 56.31 & 71.54 & 59.89 & 32.50 & 68.27 & 91.90 & 29.02 & 86.39 & 63.09 \\
Bagel~\citep{bagel}                    & 61.89 & 67.73 & 56.86 & 76.59 & 65.85 & 23.85 & 70.64 & 90.08 &  0.00 & 85.42 & 59.91 \\
BLIP3-o~\citep{chen2025blip3o}                  & 64.25 & 64.77 & 54.57 & 67.23 & 69.05 & 36.78 & 65.99 & 92.81 &  0.00 & 79.97 & 59.57 \\
Emu3~\citep{sheynin2024emu}                     & 40.21 & 50.11 & 36.21 & 43.87 & 50.67 & 19.32 & 48.60 & 87.50 &  1.15 & 76.42 & 45.42 \\
SDXL~\citep{podell2023sdxl}                     & 34.44 & 44.66 & 26.68 & 30.70 & 48.48 & 10.34 & 46.37 & 87.45 &  0.00 & 72.28 & 40.22 \\
\midrule
\multicolumn{12}{c}{\textit{Train-Time Scaling w/ RationalRewards}} \\
\midrule
FLUX.1-dev               & 62.24 & 67.20 & 45.75 & 70.84 & 62.30 & 29.77 & 66.88 & 85.00 & 32.18 & 87.50 & 60.97 \\
+MultiReward             & 59.78 & 68.23 & 44.21 & 74.37 & 59.33 & 28.25 & 68.35 & 76.05 & 36.21 & 86.03 & 60.12 \\
+Qwen3-VL-32B            & 65.47 & 72.68 & 53.28 & 71.82 & 60.78 & 33.24 & 71.85 & 85.53 & 42.15 & 89.47 & 66.53 \\
\rowcolor{yellow!20} +RationalRewards         & 67.40 & 76.36 & 57.67 & 72.15 & 60.29 & 40.53 & 74.59 & 87.20 & 52.57 & 90.61 & \textbf{70.34} \\
\midrule
SD-3.5-Medium            & 60.41 & 66.99 & 53.35 & 70.31 & 59.89 & 37.73 & 68.78 & 89.80 & 15.23 & 84.34 & 60.71 \\
\rowcolor{yellow!20} +RationalRewards         & 64.36 & 81.49 & 67.98 & 75.88 & 58.68 & 42.37 & 75.60 & 89.60 & 10.05 & 91.77 & \textbf{70.56} \\
+MultiReward             & 57.03 & 66.67 & 51.03 & 75.37 & 57.22 & 34.86 & 67.51 & 77.60 & 21.84 & 86.71 & 62.55 \\
+Qwen3-VL-32B            & 61.23 & 74.48 & 63.85 & 75.34 & 59.67 & 31.23 & 72.84 & 84.73 & 14.87 & 88.86 & 66.71 \\
\midrule
Qwen-Image               & 82.49 & 87.93 & 72.94 & 86.56 & 60.96 & 51.59 & 80.08 & 94.70 & 72.13 & 94.15 & 78.36 \\
+MultiReward             & 79.52 & 86.45 & 70.91 & 88.53 & 58.43 & 48.62 & 80.55 & 83.75 & 67.18 & 92.17 & 75.61 \\
+Qwen3-VL-32B            & 81.95 & 87.45 & 76.42 & 87.73 & 62.93 & 51.14 & 81.55 & 95.20 & 75.67 & 95.63 & 80.17 \\
\rowcolor{yellow!20} +RationalRewards         & 82.11 & 87.82 & 78.82 & 88.07 & 66.21 & 52.88 & 82.21 & 96.60 & 79.76 & 96.57 & \textbf{82.60} \\
\bottomrule
\end{tabular}%
}
\vspace{-.5cm}
\label{tab:unigenfull}
\end{table*}
}

\def\picatable{

\begin{table*}[t!]
    \centering
    \caption{\small We test OOD Generalization of \model on physics-aware editing tasks (PICABench).}
    \label{tab:picafull}
    \resizebox{\textwidth}{!}{%
    \begin{tabular}{lccccccccc}
        \toprule
        \textbf{Model} & \textbf{LightProp} & \textbf{LightSrcEff} & \textbf{Reflection} & \textbf{Refraction} & \textbf{Deformation} & \textbf{Causality} & \textbf{GlobalStateTrans} & \textbf{LocalStateTrans} & \textbf{Overall} \\
        \midrule
        Nano Banana      & 53.27 & 54.45 & 55.99 & 56.58 & 47.68 & 58.93 & 58.17 & 52.81 & 55.40 \\
        GPT-Image-1      & 59.56 & 61.99 & 52.61 & 61.84 & 44.99 & 53.16 & 70.53 & 51.56 & 57.83 \\
        Nano Banana Pro  & 59.32 & 64.69 & 61.38 & 60.09 & 53.55 & 64.70 & 72.08 & 63.41 & 63.29 \\
        Seedream 4.0     & 58.84 & 66.04 & 58.85 & 62.72 & 50.12 & 67.09 & 77.37 & 63.62 & 64.91 \\
        GPT-Image-1.5    & 62.95 & 71.43 & 61.21 & 62.28 & 57.18 & 67.23 & 76.93 & 66.11 & 67.01 \\
        \midrule
        Uniworld-V1      & 37.77 & 34.50 & 37.44 & 30.70 & 30.32 & 34.18 & 28.81 & 38.67 & 33.80 \\
        Bagel            & 54.48 & 63.34 & 55.28 & 55.70 & 42.05 & 52.32 & 68.43 & 54.05 & 56.44 \\
        Bagel-Think      & 42.86 & 52.29 & 43.17 & 48.25 & 40.10 & 38.40 & 53.86 & 46.99 & 45.91 \\
        OmniGen2         & 51.09 & 47.98 & 48.74 & 45.18 & 42.79 & 48.24 & 52.76 & 42.41 & 48.18 \\
        Step1X-Edit      & 43.10 & 52.29 & 47.05 & 47.37 & 40.34 & 46.69 & 57.95 & 47.19 & 48.83 \\
        \midrule
        Flux.1 Kontext [dev] & 48.43 & 53.64 & 43.84 & 43.86 & 33.74 & 34.04 & 41.06 & 37.01 & 41.07 \\
        +PromptEnhance   & 48.91 & 55.53 & 45.87 & 43.86 & 38.14 & 44.30 & 44.15 & 43.87 & 45.28 \\
        \rowcolor{yellow!20} +PT (\model)    & 53.27 & 56.87 & 51.43 & 40.35 & 41.08 & 43.60 & 55.30 & 43.04 & \textbf{48.12} \\
        +PICA SFT             & 49.64 & 51.21 & 47.22 & 46.49 & 33.99 & 35.44 & 39.29 & 40.75 & 41.93 \\
        \rowcolor{yellow!20} +RL (\model)              & 50.85 & 51.75 & 54.81 & 41.23 & 39.36 & 36.99 & 43.27 & 35.76 & 44.25 \\
        \midrule
        Qwen-Image-Edit  & 52.54 & 52.02 & 49.07 & 57.46 & 38.14 & 42.62 & 57.73 & 47.82 & 49.71 \\
        +PromptEnhance   & 54.24 & 58.49 & 50.42 & 49.12 & 42.30 & 43.46 & 57.40 & 50.31 & 50.97 \\
        \rowcolor{yellow!20} +PT (\model)    & 61.26 & 63.34 & 61.55 & 55.70 & 43.28 & 46.27 & 57.28 & 56.55 & \textbf{55.65} \\
        +PICA SFT             & 52.89 & 60.47 & 55.19 & 56.12 & 40.99 & 46.24 & 55.25 & 51.27 & 52.06 \\
        \rowcolor{yellow!20} +RL (\model)              & 59.56 & 63.07 & 60.71 & 55.26 & 41.32 & 45.85 & 56.40 & 49.69 & 54.11 \\
        \bottomrule
    \end{tabular}
    }
\end{table*}
}

\def\edittable{
\begin{table*}[t!]
\centering
\caption{\small We perform RL tuning and test-time prompt tuning to test \model on image editing. On ImgEdit-Bench and GEdit-Bench-EN,  trading test-time evaluation for better generation yields surprising gains.}
\resizebox{\textwidth}{!}{%
\begin{tabular}{lcccccccccc|ccc}
\toprule
\multirow{2}{*}{\textbf{Model}} & \multicolumn{10}{c}{\textbf{ImgEdit-Bench}} & \multicolumn{3}{|c}{\textbf{GEdit-Bench-EN}} \\
\cmidrule(lr){2-11} \cmidrule(lr){12-14}
 & \textbf{Add} & \textbf{Adjust} & \textbf{Extract} & \textbf{Replace} & \textbf{Remove} & \textbf{Background} & \textbf{Style} & \textbf{Compose} & \textbf{Action} & \textbf{Overall} & \textbf{G\_SC} & \textbf{G\_PQ} & \textbf{G\_O} \\
\midrule
AnyEdit~\citep{yu2025anyedit} & 3.18 & 2.95 & 1.88 & 2.47 & 2.23 & 2.23 & 2.85 & 1.56 & 2.65 & 2.45 & 3.18 & 5.82 & 3.21 \\
UltraEdit~\citep{zhao2024ultraedit} & 3.44 & 2.81 & 2.13 & 2.96 & 1.45 & 2.86 & 3.76 & 1.91 & 2.98 & 2.70 & - & - & - \\
Step1X-Edit~\citep{step1edit} & 3.88 & 3.14 & 1.76 & 3.40 & 2.41 & 3.16 & 4.63 & 2.64 & 2.52 & 3.06 & 7.66 & 7.35 & 6.97 \\
BAGEL~\citep{bagel} & 3.56 & 3.31 & 1.70 & 3.30 & 2.62 & 3.24 & 4.49 & 2.38 & 4.17 & 3.20 & 7.36 & 6.83 & 6.52 \\
OmniGen2~\citep{omnigen} & 3.57 & 3.06 & 1.77 & 3.74 & 3.20 & 3.57 & 4.81 & 2.52 & 4.68 & 3.44 & 7.16 & 6.77 & 6.41 \\
Ovis-U1~\citep{ovis} & 4.13 & 3.62 & 2.98 & 4.45 & 4.06 & 4.22 & 4.69 & 3.45 & 4.61 & 4.00 & - & - & 6.42 \\
GPT-Image-1~\citep{gptimage} & 4.61 & 4.33 & 2.90 & 4.35 & 3.66 & 4.57 & 4.93 & 3.96 & 4.89 & 4.20 & 7.85 & 7.62 & 7.53 \\
\midrule
\multicolumn{14}{c}{\textit{Train/Test Time Scaling /w \model}} \\
\midrule
Flux.1 Kontext [dev] & 3.76 & 3.45 & 2.15 & 3.98 & 2.94 & 3.78 & 4.38 & 2.96 & 4.26 & 3.52 & 7.16 & 7.37 & 6.51 \\
+RL (EditReward) & 3.91 & 3.83 & 2.39 & 4.15 & 2.99 & 3.99 & 4.56 & 2.73 & 4.11 & 3.66 & 7.38 & 7.53 & 6.88 \\
+RL (Qwen3-VL-32B) & 3.95 & 3.90 & 2.41 & 4.12 & 2.95 & 3.96 & 4.45 & 2.82 & 4.30 & 3.67 & 7.42 & 7.48 & 6.82 \\
\rowcolor{yellow!20}  +RL (\model) & 4.21 & 4.34 & 2.68 & 4.33 & 2.92 & 4.05 & 4.37 & 3.09 & 4.41 & 3.84 & 7.75 & 8.24 & \textbf{7.37} \\
+PT (\model)  & 3.96 & 4.16 & 3.37 & 4.38 & 3.84 & 4.12 & 4.55 & 2.70 & 4.29 & \textbf{4.01} & 7.77 & 7.61 & 7.23 \\
\midrule
Qwen-Image-Edit & 4.38 & 4.16 & 3.43 & 4.66 & 4.14 & 4.38 & 4.81 & 3.18 & 4.69 & 4.27 & 8.00 & 7.86 & 7.56 \\
+RL (EditReward) & 4.34 & 4.22 & 3.87 & 4.67 & 4.18 & 4.20 & 4.83 & 3.36 & 4.54 & 4.25 & 8.36 & 7.91 & 7.77 \\
+RL (Qwen3-VL-32B) & 4.40 & 4.18 & 3.35 & 4.60 & 4.10 & 4.35 & 4.80 & 3.10 & 4.72 & 4.25 & 8.42 & 7.83 & 7.79 \\
\rowcolor{yellow!20}  +RL (\model) & 4.41 & 4.32 & 4.09 & 4.63 & 4.26 & 4.25 & 4.91 & 3.44 & 4.52 & 4.38 & 8.74 & 8.43 & 8.29 \\
+ PT (\model)  & 4.46 & 4.40 & 4.18 & 4.63 & 4.27 & 4.40 & 4.88 & 3.27 & 4.54 & \textbf{4.43} & 8.94 & 8.20 & \textbf{8.33} \\
\bottomrule
\end{tabular}%
}
\vspace{-.5cm}
\label{tab:editfull}
\end{table*}
}
\def\editablationtable{
    \begin{table*}[t!]
        \centering
        \caption{\small Ablation of \model as dual-space optimizer on editing tasks. For prompt space tuning, we compare pre-generation PromptEnhance~\citep{promptenhancer}. For parameter space tuning, we compare SFT and RL with different rewards. We include OOD physics-aware editing, PICA-Bench with representative aspects (Left), and generic editing benchmarks (Right).}
        \label{tab:ablation_sidebyside}
        \begin{minipage}[t]{0.51\textwidth}
        \centering
        \label{tab:ablation_pica}
        
        \resizebox{\textwidth}{!}{%
        \begin{tabular}{lccccc}
            \toprule
            \multirow{2}{*}{\textbf{Model}} & \multicolumn{3}{c}{\textbf{Representative Aspects}} & \multirow{2}{*}{\textbf{Overall}} \\
            \cmidrule(lr){2-4}
             & \textbf{Light} & \textbf{Reflec.} & \textbf{Deform.} & \\
            \midrule
            Flux.1 Kontext [dev] & 53.64 & 43.84 & 33.74 & 41.07 \\
            +PromptEnhance   & 55.53 & 45.87 & 38.14 & 45.28 \\
            \rowcolor{yellow!20} +PT (\model)    & 56.87 & 51.43 & 41.08 & \textbf{48.12} \\
            +PICA SFT        & 51.21 & 47.22 & 33.99 & 41.93 \\
            \rowcolor{yellow!20} +RL (\model)     & 51.75 & 54.81 & 39.36 & 44.25 \\
            \midrule
            Qwen-Image-Edit  & 52.02 & 49.07 & 38.14 & 49.71 \\
            +PromptEnhance   & 58.49 & 50.42 & 42.30 & 50.97 \\
            \rowcolor{yellow!20} +PT (\model)    & 63.34 & 61.55 & 43.28 & \textbf{55.65} \\
            +PICA SFT        & 60.47 & 55.19 & 40.99 & 52.06 \\
            \rowcolor{yellow!20} +RL (\model)     & 63.07 & 60.71 & 41.32 & 54.11 \\
            \bottomrule
        \end{tabular}%
        }
        \end{minipage}%
        \hfill
        \begin{minipage}[t]{0.48\textwidth}
        \centering
        \label{tab:ablation_edit}
        
        \resizebox{\textwidth}{!}{%
        \begin{tabular}{lc|ccc}
        \toprule
        \multirow{2}{*}{\textbf{Model}} & \textbf{ImgEdit} & \multicolumn{3}{c}{\textbf{GEdit-Bench-EN}} \\
        \cmidrule(lr){2-2} \cmidrule(lr){3-5}
         & \textbf{Overall} & \textbf{G\_SC} & \textbf{G\_PQ} & \textbf{G\_O} \\
        \midrule
        Flux.1 Kontext [dev] & 3.52 & 7.16 & 7.37 & 6.51 \\
        +RL (EditReward) & 3.66 & 7.38 & 7.53 & 6.88 \\
        +RL (Qwen3-VL-32B) & 3.67 & 7.42 & 7.48 & 6.82 \\
        
        \rowcolor{yellow!20} +RL (\model) & 3.84 & 7.75 & 8.24 & \textbf{7.37} \\
        \rowcolor{yellow!20} +PT (\model) & \textbf{4.01} & 7.77 & 7.61 & 7.23 \\
        \midrule
        Qwen-Image-Edit & 4.27 & 8.00 & 7.86 & 7.56 \\
        +RL (EditReward) & 4.25 & 8.36 & 7.91 & 7.77 \\
        +RL (Qwen3-VL-32B) & 4.25 & 8.42 & 7.83 & 7.79 \\
        
        \rowcolor{yellow!20} +RL (\model) & 4.38 & 8.74 & 8.43 & 8.29 \\
        \rowcolor{yellow!20} +PT (\model) & \textbf{4.43} & 8.94 & 8.20 & \textbf{8.33} \\
        \bottomrule
        \end{tabular}%
        }
        \end{minipage}
        \end{table*}
}
\def\inferencefig{
\begin{figure*}[t!]
    \vspace{-0.3cm}
    \centering
\includegraphics[width=0.99\linewidth]{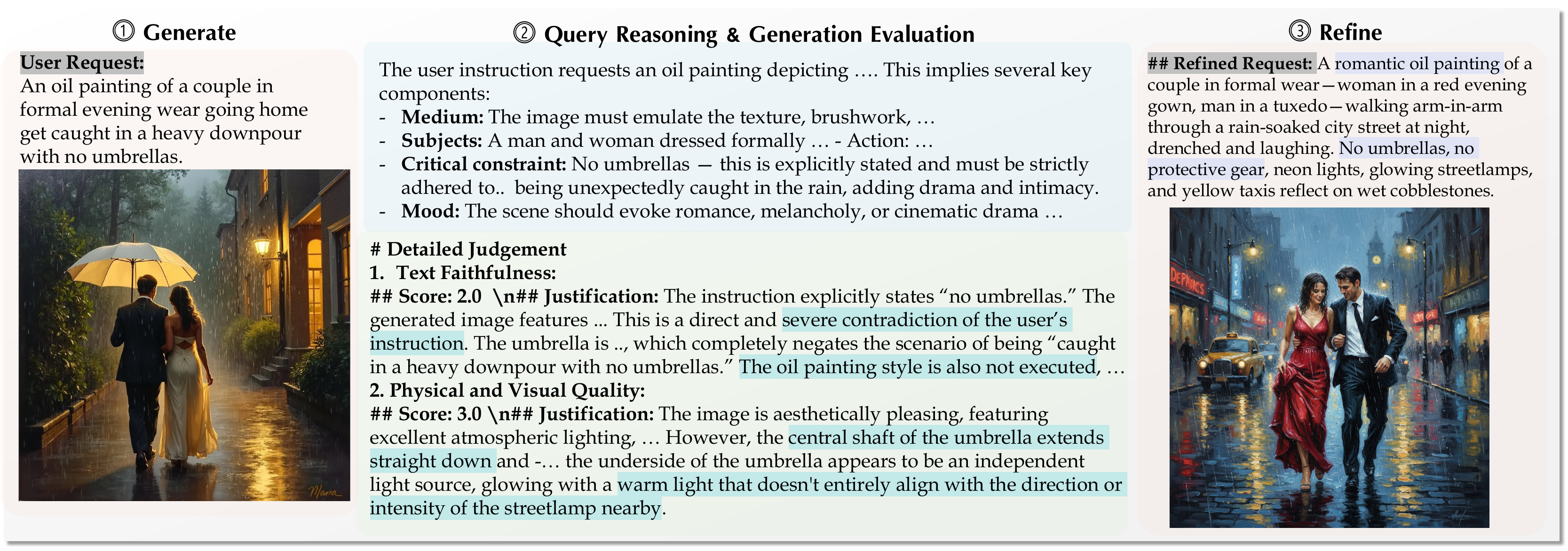}

    \caption{\small   Test-Time Prompt Refinement via ``Generate-Critique-Refine'' loop with \model.   }
    \label{fig_inference}
\end{figure*}
}

\def\rmqualityfigure{
\begin{figure*}[!t]
    \centering
\includegraphics[width=\linewidth]{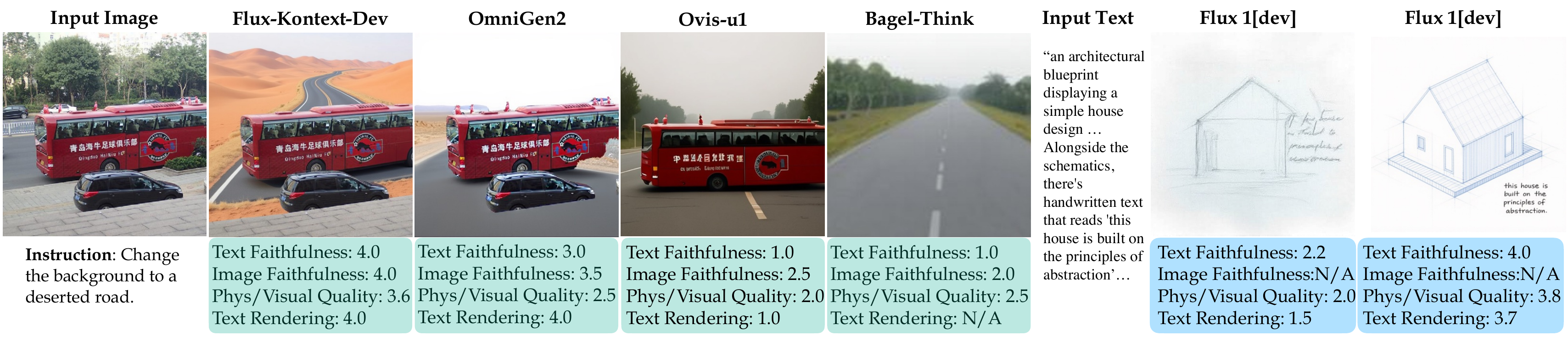}
    \caption{\small Example pointwise scores rated by \model for image/text-to-image generations (rationales omitted). \model evaluates each result across multiple dimensions.}
    \vspace{-.3cm}
    \label{fig_examples_scores}
\end{figure*}
}
\def\qualityfigure{
\begin{figure*}[t]
    \centering
\includegraphics[width=\linewidth]{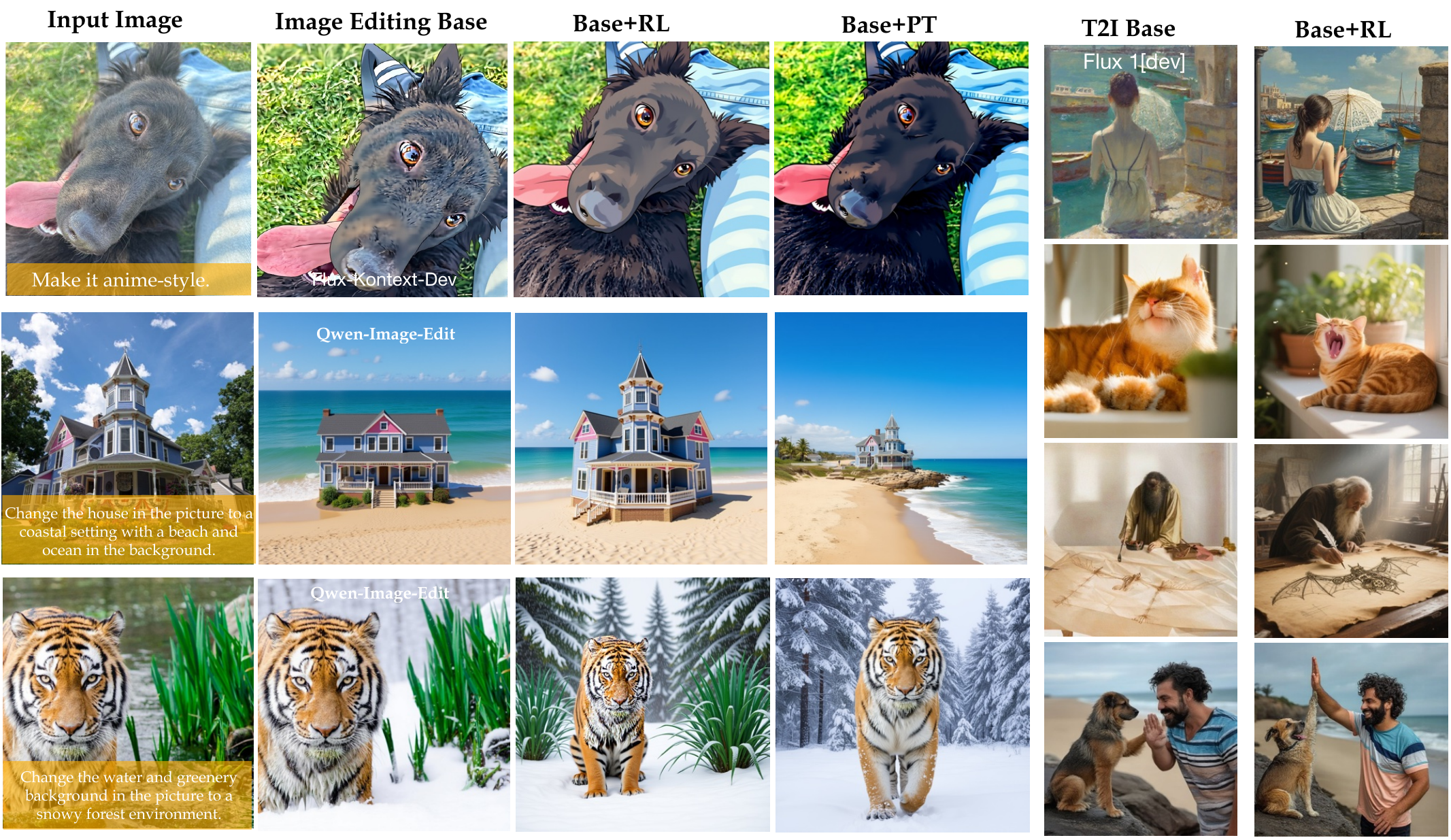}
    \caption{\small Qualitative results on image/text-to-image tasks optimized with reinforcement learning (RL) and prompt tuning (PT) using \model.}
    \label{fig_before_after_rl_pt}
\end{figure*}
}

%% file: prompt_template.tex


\tcbset{
    promptbox/.style={
        colback=blue!3!white,
        colframe=blue!25!black,
        fonttitle=\bfseries\small,
        coltitle=white,                        
        colbacktitle=blue!20!gray!80!black,    
        sharp corners=south,
        rounded corners=north,
        boxrule=0.8pt,
        fontupper=\small,
        boxsep=3pt,
        left=4pt, right=4pt, top=6pt, bottom=3pt,
    }
}

\def\promptpair{
\begin{tcolorbox}[promptbox, title={\color{white}(a) Pairwise Rationale Generation (Phase 1)}]
\textbf{Input:} Instruction \texttt{\{inst\}}, Source Image, Edited Image A, Edited Image B\\[1pt]
\textbf{Task:} Compare two edited images according to the instruction.\\[2pt]
\textbf{Aspects} (each scored 1--4): \textit{Text Faithfulness} $\cdot$ \textit{Image Faithfulness} $\cdot$ \textit{Physical \& Visual Quality} $\cdot$ \textit{Text Rendering}\\[2pt]
\textbf{Preference Anchor:} ``Hint: human preference is: \texttt{\{label\}}''\\[2pt]
\textbf{Output Format:}\\[-2pt]
{\footnotesize
\texttt{[Understanding of the user request]}\\
\texttt{\# Detailed Judgement}\\
\texttt{1. Text Faithfulness:}\\
\texttt{~~\#\# Justification: [...] \#\# Score A: [...] \#\# Score B: [...] \#\# Winner: [...]}\\
\texttt{2--4. \textit{(same structure for remaining aspects)}}\\
\texttt{\# Summary: [...]}
}
\end{tcolorbox}
}  
\def\promptpoint{}  

\newcommand{\PromptTemplateFigure}{
\begin{figure}[t!]
\centering
\promptpair



\caption{\small Prompt templates for rationale generation. Template for pointwise projection adopts similar form but conditions on the validated pairwise rationales as hints. Full prompts with detailed rubrics are in the appendix.}
\label{fig_pairwise_prompt}
\end{figure}
}

%% file: appendix.tex
\newpage\appendix

\section{Extended Experimental Results}
\label{app:results}

\paragraph{Full Text-to-Image Results on UniGenBench++}
\label{app:unigen-full}

Table~\ref{tab:unigenfull} provides the complete UniGenBench++ results across all categories and model variants.

\unigentable

\paragraph{Full Image Editing Results}
\label{app:edit-full}

Table~\ref{tab:editfull} provides the complete results on generic image editing benchmarks.

\edittable

\paragraph{Full PICA-Bench Results}
\label{app:pica-full}

Table~\ref{tab:picafull} provides the complete PICA-Bench results across all physics-aware aspects, extending the representative results shown in Table~3 (left panel) of the main text.

\picatable

\paragraph{Training Curves and Visualizations}
\label{app:training-curves}

\begin{figure}[t]
    \centering
    \includegraphics[width=\linewidth]{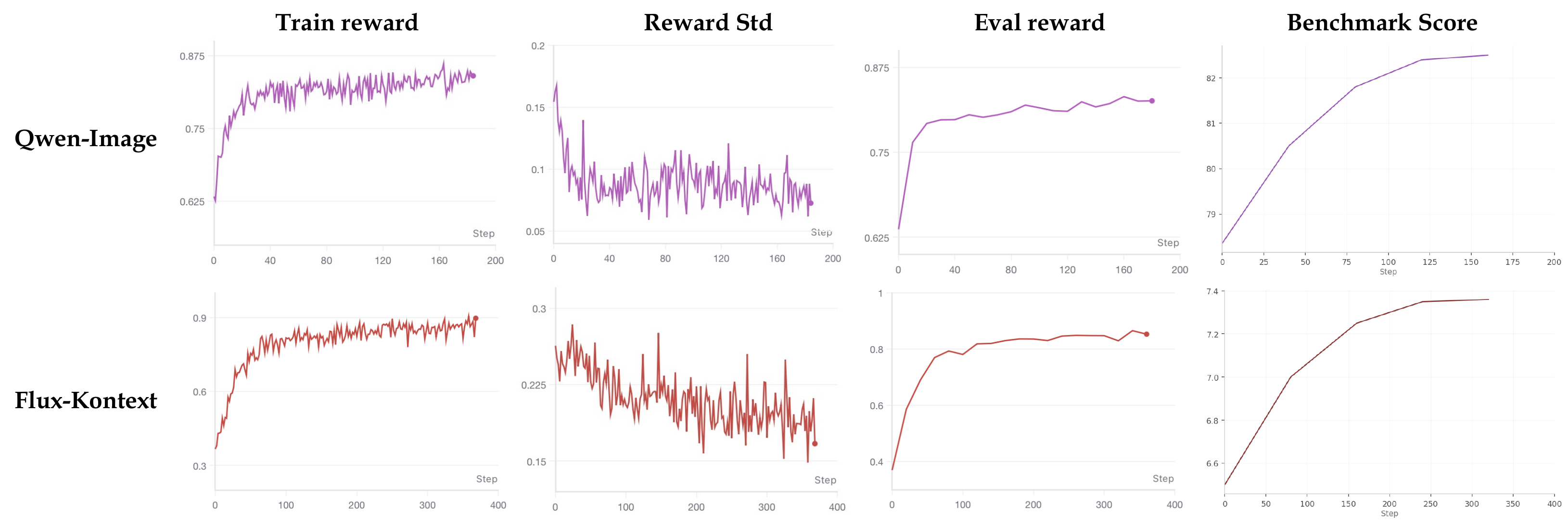}
    \caption{
    RL with \model on Qwen-Image (text-to-image generator) and Flux-Kontext [dev] (image-to-image editing). The reward standard-deviation gradually decays as training proceeds. Crucially, the evaluation reward curve on held-out eval-set align well with the score curve on target test benchmarks. }
    \label{fig:rl_curve}
\end{figure}

\begin{figure}[t!]
    \centering
    \includegraphics[width=1.0\linewidth]{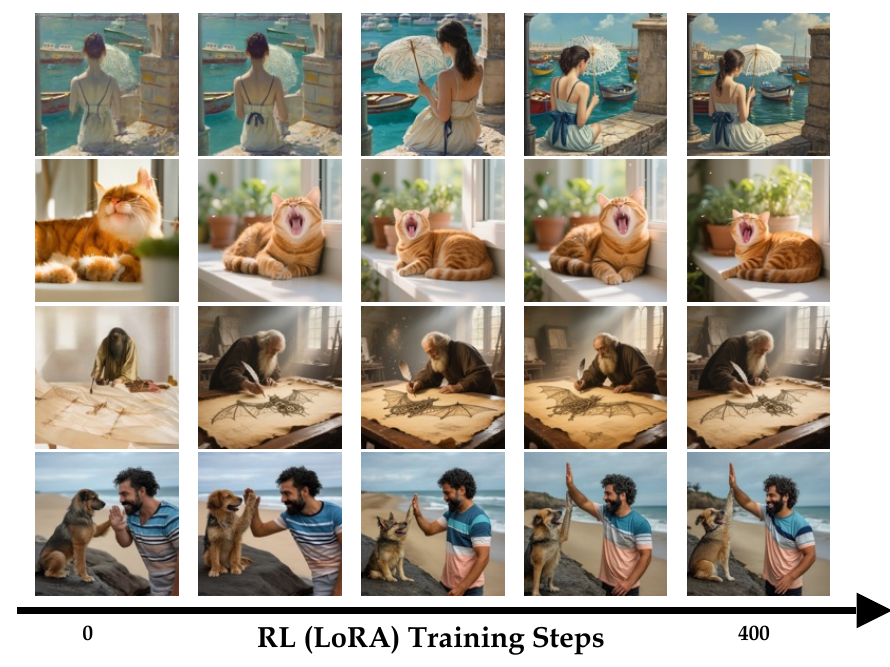}
    \caption{The evolution of generation quality of RL using \model}
    \label{fig:t2i_rl_evolve}
\end{figure}

This section provides training curves referenced in Section~3.2 of the main text, demonstrating that RationalRewards provides stable reward gradients with reduced reward hacking, as shown in Fig.~\ref{fig:rl_curve}. Qualitative Results throughout RL training are visualized in Fig.~\ref{fig:t2i_rl_evolve}. 

\begin{figure*}[t]
    \centering
    \includegraphics[width=\textwidth]{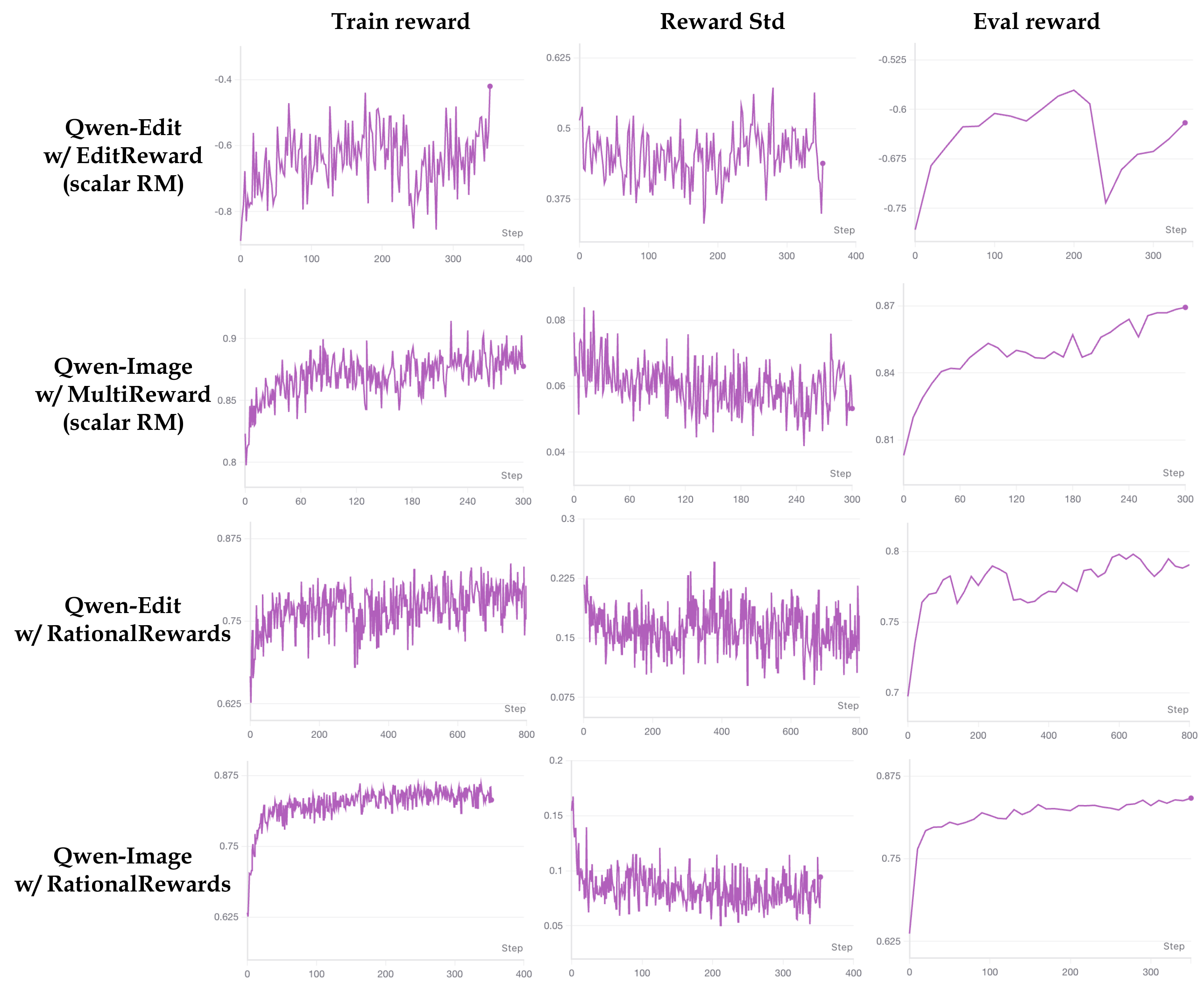}
    \caption{Training curves comparison between \model and scalar reward model, EditReward~\citep{editreward} and MultiReward used in DiffusionNFT~\cite{diffusionnft}.  }
    \label{fig_curve_compare}
\end{figure*}

\begin{figure*}[t]
    \centering
    \includegraphics[width=\textwidth]{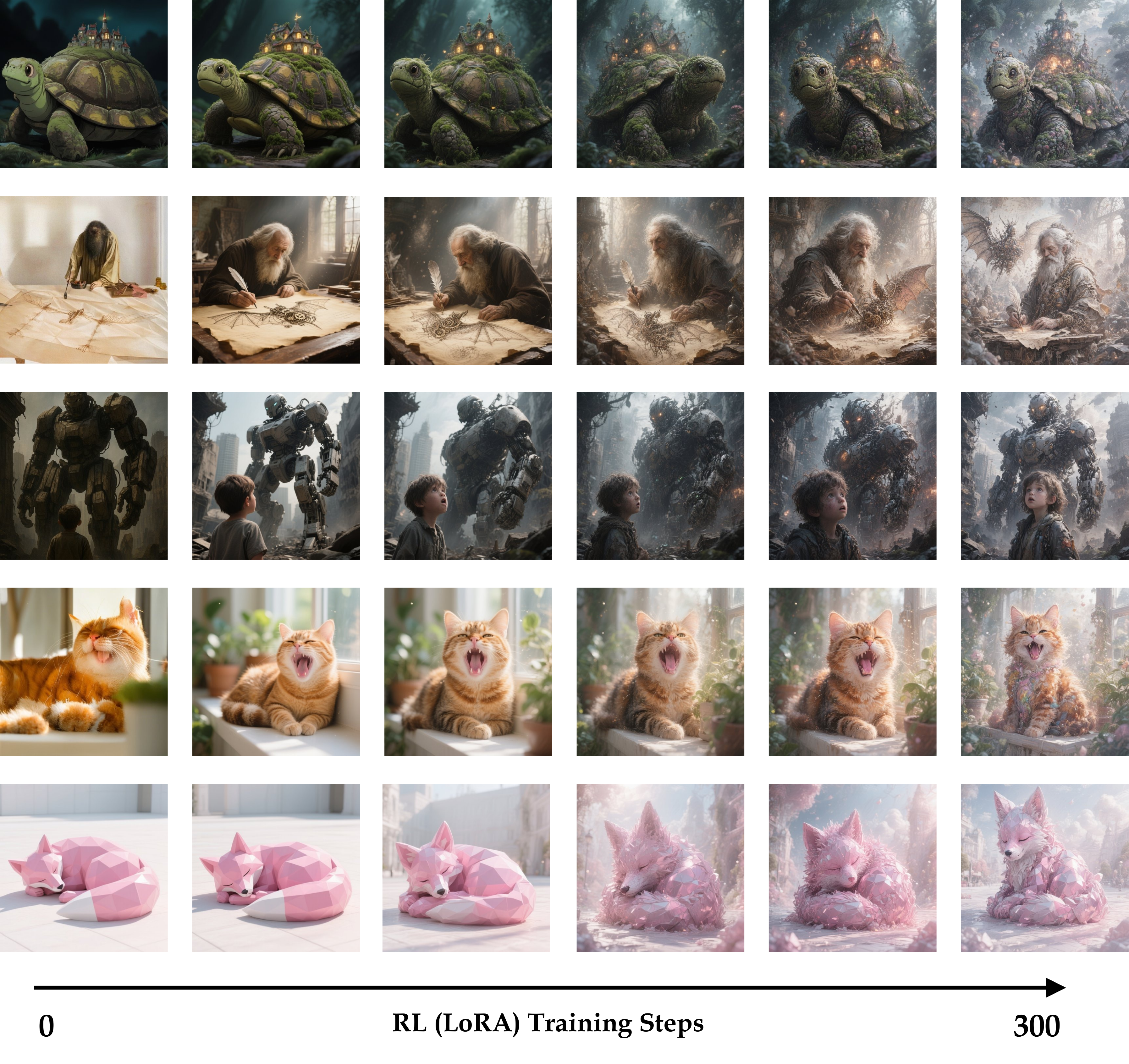}
    \caption{Text-to-Image RL using scalar reward model demonstrates reward hacking -- while the reward increases, the visual quality of generations degrades notably. }
    \label{fig_reward_hacking}
\end{figure*}

\begin{figure*}[t]
    \centering
\includegraphics[width=\linewidth]{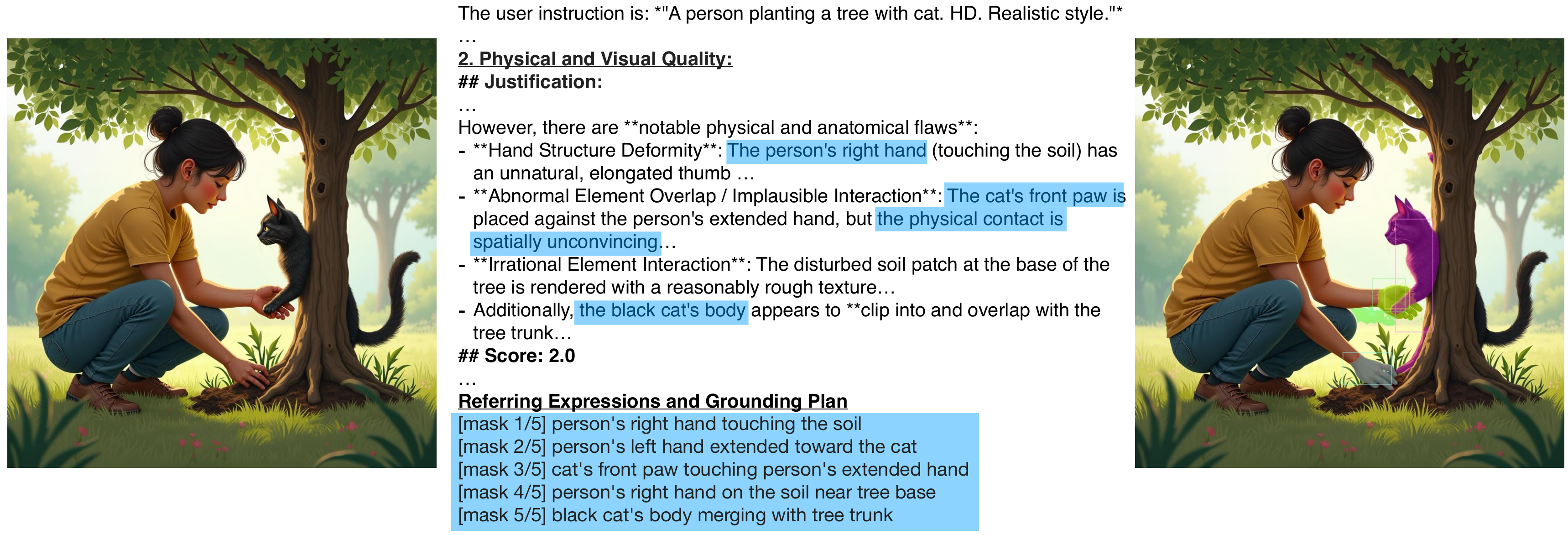}
\caption{\small Illustration of Critique Visualization.\model first analyzes the image and provides critique rationales, then summarizes them and generates referring expressions for GroundingDINO and SAM to produce segmentation masks for problematic regions.}
    \label{fig_critique_visual}
    \end{figure*}

\paragraph{Reward Hacking and Visualizations.} Fig.~\ref{fig_curve_compare} compares \model with representative scalar reward models used in text-to-image and image-to-image generation RL. \model demonstrates nice properties of smooth, converging reward curve and standard-deviation curve. In contrast, EditReward remains high variances, leading to unstable reward curve. MultiReward exhibits low variances because it does not suffice to differantiate generations of high-capability generators. Fig.~\ref{fig_reward_hacking} shows clear visual evidence of reward hacking. 
\paragraph{Critique Visualization.}
We provide additional example use case of \model, which visualizes problematic regions and grounds its scoring in the image.
Specifically, \model is further fine-tuned to generate structured referring expressions that describe problematic regions. 
These expressions are used by GroundingDINO to localize the regions, and the resulting bounding boxes are then used by SAM to produce segmentation masks as show in~\autoref{fig_critique_visual}.

\section{ELBO Derivation and Theoretical Details}
\label{app:elbo}

This appendix provides the complete derivation of the Evidence Lower Bound (ELBO) presented in Eq.~1 of the main text (Section~2.1) and discusses the theoretical assumptions underlying the pointwise projection strategy.

\subsection{Full ELBO Derivation}
\label{app:elbo-derivation}

We begin from the log marginal likelihood of the observed preference $y$ given input $x = (I_A, I_B, c)$, where $I_A, I_B$ are two generated images and $c$ is the conditioning user request. We introduce a latent natural language rationale $z$ that explains the preference:
\begin{align}
\log P_\theta(y \mid x) &= \log \int P_\theta(y, z \mid x) \, dz.
\end{align}

Since this marginal is intractable (the integral is over all possible natural-language rationales), we introduce a variational distribution $q_\phi(z \mid x, y)$---the \emph{posterior} over rationales given both the input and the known preference. Multiplying and dividing inside the integral:
\begin{align}
\log P_\theta(y \mid x) &= \log \int q_\phi(z \mid x, y) \frac{P_\theta(y, z \mid x)}{q_\phi(z \mid x, y)} \, dz = \log \, \mathbb{E}_{z \sim q_\phi(\cdot \mid x, y)} \left[ \frac{P_\theta(y, z \mid x)}{q_\phi(z \mid x, y)} \right].
\end{align}

Applying Jensen's inequality ($\log \mathbb{E}[\cdot] \geq \mathbb{E}[\log \cdot]$, since $\log$ is concave):
\begin{align}
\log P_\theta(y \mid x) &\geq \mathbb{E}_{z \sim q_\phi} \left[ \log \frac{P_\theta(y, z \mid x)}{q_\phi(z \mid x, y)} \right] \equiv \mathcal{L}_{\text{ELBO}}.
\end{align}

We now decompose the joint $P_\theta(y, z \mid x)$ using the chain rule $P_\theta(y, z \mid x) = P_\theta(y \mid x, z) \cdot P_\theta(z \mid x)$:
\begin{align}
\mathcal{L}_{\text{ELBO}} &= \mathbb{E}_{z \sim q_\phi} \left[ \log P_\theta(y \mid x, z) + \log P_\theta(z \mid x) - \log q_\phi(z \mid x, y) \right] \\
&= \underbrace{\mathbb{E}_{z \sim q_\phi} \left[ \log P_\theta(y \mid x, z) \right]}_{\text{Term 1: Prediction}} + \mathbb{E}_{z \sim q_\phi} \left[ \log \frac{P_\theta(z \mid x)}{q_\phi(z \mid x, y)} \right] \\
&= \underbrace{\mathbb{E}_{z \sim q_\phi} \left[ \log P_\theta(y \mid x, z) \right]}_{\text{Term 1: Prediction}} - \underbrace{D_{\mathrm{KL}}\!\left( q_\phi(z \mid x, y) \,\|\, P_\theta(z \mid x) \right)}_{\text{Term 2: Regularization}},
\end{align}
which yields Eq.~1 in the main text.

\paragraph{Tightness of the Bound.} The gap between the ELBO and the true log-likelihood is given exactly by the KL divergence between the variational posterior and the true posterior:
\begin{align}
\log P_\theta(y \mid x) = \mathcal{L}_{\text{ELBO}} + D_{\mathrm{KL}}\!\left( q_\phi(z \mid x, y) \,\|\, P_\theta(z \mid x, y) \right).
\end{align}
This follows directly from the definition of KL divergence:
\begin{align}
D_{\mathrm{KL}}\!\left( q_\phi \,\|\, P_\theta(\cdot \mid x, y) \right) &= \mathbb{E}_{q_\phi} \left[ \log \frac{q_\phi(z \mid x, y)}{P_\theta(z \mid x, y)} \right] \\
&= \mathbb{E}_{q_\phi} \left[ \log q_\phi(z \mid x, y) - \log P_\theta(z, y \mid x) + \log P_\theta(y \mid x) \right] \\
&= -\mathcal{L}_{\text{ELBO}} + \log P_\theta(y \mid x).
\end{align}
Since $D_{\mathrm{KL}} \geq 0$ and $\log P_\theta(y \mid x)$ is fixed with respect to $\phi$, maximizing the ELBO is equivalent to minimizing the KL divergence between the variational posterior $q_\phi(z \mid x, y)$ and the true posterior $P_\theta(z \mid x, y)$.

\paragraph{Mapping ELBO Terms to Pipeline Phases.} The three terms of the decomposition correspond directly to the three phases of the PARROT pipeline (Figure~3):
\begin{enumerate}[leftmargin=*]
    \item \textbf{Phase 1 (Rationale Generation)} constructs the variational posterior $q_\phi(z \mid x, y)$ by prompting a teacher VLM with preference-anchored instructions. The preference label $y$ is provided as a hint, focusing generation on rationales consistent with the observed preference.
    \item \textbf{Phase 2 (Consistency Filtering)} maximizes Term 1, $\mathbb{E}_{q_\phi}[\log P_\theta(y \mid x, z)]$, by retaining only rationales $z$ for which the preference $y$ can be recovered from $(x, z)$ alone (Eq.~2). This restricts $q_\phi$'s effective support to the high-likelihood region, ensuring predictive sufficiency.
    \item \textbf{Phase 3 (Foresight Distillation)} minimizes Term 2, $D_{\mathrm{KL}}(q_\phi(z \mid x, y) \| P_\theta(z \mid x))$, by training the student model $P_\theta(z \mid x)$ to generate rationales without access to $y$. Since $q_\phi$ is fixed, this reduces to maximizing $\mathbb{E}_{q_\phi}[\log P_\theta(z \mid x)]$, which is precisely the standard supervised fine-tuning (SFT) objective on the filtered posterior samples.
\end{enumerate}

\paragraph{Factorization Assumption.} The derivation assumes the joint factorizes as $P_\theta(z, y \mid x) = P_\theta(y \mid x, z) \cdot P_\theta(z \mid x)$, i.e., the model first generates a rationale $z$ given the input $x$, then predicts the preference $y$ conditioned on both. This autoregressive factorization is natural for language models, where $z$ (the rationale) is generated token-by-token before the preference prediction $y$. The factorization encodes the causal assumption that the rationale mediates the preference judgment---the model must ``show its work'' before committing to a decision~\citep{wang2025illusion, wang2020learning}.

\subsection{Justification for Pointwise Projection}
\label{app:pointwise-justification}

The pointwise projection strategy (Section~2.1, main text) extends the pairwise ELBO framework to absolute scoring of individual images. We discuss the assumptions underlying this extension.

\paragraph{Shared Evaluation Principles.} The core assumption is that the evaluation criteria underlying pairwise preference (e.g., ``Image A has better text faithfulness than Image B because...'') are transferable to absolute assessment (e.g., ``This image has a text faithfulness score of 3.2 because...''). This is grounded in the observation that the same rubric dimensions---text faithfulness, image faithfulness, physical quality, and text rendering---apply in both settings, differing only in whether the assessment is relative or absolute.

\paragraph{Role of Pairwise Rationales as Reference Hints.} During pointwise projection, the validated pairwise rationale $z_{\text{pair}}$ serves as a reference hint to guide the teacher's attention toward specific defects or qualities already identified in the pairwise comparison. This anchoring reduces the variance of pointwise assessments by providing concrete evidence (e.g., ``as noted in the comparison, the text rendering in this image has minor misspellings'') rather than requiring the teacher to identify all issues from scratch. The quality of pointwise rationales thus inherits from the ELBO-filtered pairwise rationales.

\paragraph{Potential Failure Modes.} We acknowledge two potential failure modes: (1) \emph{calibration drift}, where the relative ranking between two images is correct but the absolute scores are miscalibrated (e.g., both images receive high scores despite one being clearly inferior); and (2) \emph{context dependence}, where the teacher's absolute assessment is influenced by the identity of the comparison partner in the pairwise rationale, rather than being truly absolute. We mitigate (1) through float-valued scoring with detailed rubric anchors (Appendix~\ref{app:rubrics}) and (2) by instructing the teacher to assess ``as if by your own judgement'' independently of the reference hint.

\section{Prompt Templates}
\label{app:prompts}

This appendix provides the complete prompt templates used across all phases of the PARROT pipeline and the Generate--Critique--Refine (GCR) loop, referenced in Section~2.1 of the main text.

\subsection{Phase 1: Pairwise Rationale Generation Prompt}
\label{app:prompt-pairwise}

The following prompt is used to query the teacher VLM (Qwen3-VL-32B-Instruct) for pairwise rationale generation with preference anchoring. The main text (Section~2.1) shows an abbreviated version; below is the complete template.

\begin{tcolorbox}[title=Pairwise Rationale Generation --- Image Editing Variant, fonttitle=\bfseries, colback=white, colframe=black, breakable]
\small
\texttt{User Instruction: \{instruction\}} \\

\texttt{You are provided with three images:}
\begin{enumerate}[leftmargin=*, noitemsep, topsep=0pt]
    \item \texttt{The Source Image (First image)}
    \item \texttt{Edited Image A (Second image)}
    \item \texttt{Edited Image B (Third image)}
\end{enumerate}

\texttt{Your task is to compare two edited images (Edited Image A and Edited Image B) against the Source Image and the User Instruction.} \\

\texttt{To do this, you must assess each image on four critical aspects, provide justifications and scores in 1--4 scale, and determine which image is better for each aspect.} \\

\texttt{About the scores: you should try to give \textbf{float scores}. For example, float values are important to reflect fine-grained preferences when you compare two edited images.} \\

\texttt{\textbf{\#\#\# Critical Aspects \& Scoring Rubric}} \\

\texttt{\textbf{1. Text Faithfulness} (How accurately does the output follow the instruction?)} \\
\texttt{- \textbf{4 (Full match):} All key elements (objects, colors, actions) are represented exactly as described. No hallucinations or unrequested changes.} \\
\texttt{- \textbf{3 (Minor mismatch):} Most key elements are present, but minor details are missing, incorrect, or slightly inaccurate.} \\
\texttt{- \textbf{2 (Some mismatch):} Some key elements are missing, altered, or interpreted incorrectly.} \\
\texttt{- \textbf{1 (Major deviations):} Key elements are completely missing, altered, or contradicted. Instruction is ignored.} \\

\texttt{\textbf{2. Image Faithfulness} (How well are the non-edited parts and key input elements preserved?)} \\
\texttt{- \textbf{4 (Uses input fully):} All relevant elements from the input are accurately preserved or transformed as instructed.} \\
\texttt{- \textbf{3 (Minor mismatch):} Most relevant elements are preserved, but a few aspects are missing or incorrectly handled.} \\
\texttt{- \textbf{2 (Partial mismatch):} Some elements are carried over, but key aspects of the original image are lost or distorted.} \\
\texttt{- \textbf{1 (Fails to use input):} Key elements of the input image are ignored, misinterpreted, or destroyed.} \\

\texttt{\textbf{3. Physical and Visual Quality} (Technical errors, composition, realism, and physics)} \\
\texttt{- \textbf{4 (No noticeable flaws):} The image is physically plausible. No visible artifacts.} \\
\texttt{- \textbf{3 (Minor flaws):} Small inaccuracies that are noticeable but not strongly disruptive.} \\
\texttt{- \textbf{2 (Some flaws):} Clear physical or visual errors that disrupt the image.} \\
\texttt{- \textbf{1 (Severe flaws):} Major physical/visual errors.} \\

\texttt{\textbf{4. Text Rendering} (Only if the instruction involves generating text)} \\
\texttt{- \textbf{4 (Full match):} Text is correct, legible, and integrated well.} \\
\texttt{- \textbf{3 (Mostly match):} Minor misspellings or inconsistent capitalization.} \\
\texttt{- \textbf{2 (Partial match):} Major misspellings or distorted text.} \\
\texttt{- \textbf{1 (Major deviations):} Text is unreadable, severely distorted, or missing. (Use N/A if no text generation is required).} \\

\texttt{\textbf{Hint: human preference is: \{label\}}} \\

\texttt{Output your evaluation in the following format:} \\
\texttt{[ Understanding of the user request ]} \\
\texttt{\textbf{\# Detailed Judgement}} \\
\texttt{\textbf{1. Text Faithfulness:}} \\
\texttt{\textbf{\#\# Justification:} [Detailed comparison]} \\
\texttt{\textbf{\#\# Score A:} [float] \textbf{\#\# Score B:} [float] \textbf{\#\# Winner:} [A/B/Tie]} \\
\texttt{2--4. (same structure for remaining aspects)} \\
\texttt{\textbf{\# Summary:} [Overall comparison summary]}
\end{tcolorbox}

\paragraph{Text-to-Image Variant.} For text-to-image generation, the prompt is modified as follows: (1) only two images are provided (Generated Image A and Generated Image B) without a source image; (2) the ``Image Faithfulness'' dimension is replaced with N/A since there is no source image to preserve; and (3) the task description is adjusted to ``compare two generated images against the User Instruction.''

\subsection{Phase 2: Consistency Check Prompt}
\label{app:prompt-consistency}

The following prompt is used to re-query the teacher VLM \emph{without} the preference label to verify that the generated rationale $z$ alone suffices to recover the preference $y$ (Eq.~2).

\begin{tcolorbox}[title=Consistency Check Prompt, fonttitle=\bfseries, colback=white, colframe=black, breakable]
\small
\texttt{User Instruction: \{instruction\}} \\

\texttt{You are provided with the following evaluation of two edited images:} \\

\texttt{\{rationale\_z\}} \\

\texttt{Based on the above evaluation, which image is preferred overall?} \\

\texttt{Answer with ONLY one of the following:} \\
\texttt{- ``A is preferred''} \\
\texttt{- ``B is preferred''} \\
\texttt{- ``Tie''}
\end{tcolorbox}

The rationale $z$ is presented in its entirety (including per-dimension justifications, scores, and summary). The teacher must predict the preference from the rationale alone. If the predicted preference matches the ground-truth label $y$, the sample passes the consistency check ($C=1$ in Eq.~2).

\subsection{Pointwise Projection Prompt}
\label{app:prompt-pointwise}

The following prompt is used to obtain pointwise (absolute) assessments from the teacher VLM, guided by the validated pairwise rationale as a reference hint.

\begin{tcolorbox}[title=Pointwise Projection (Part 1), fonttitle=\bfseries, colback=white, colframe=black, breakable]
\small
\texttt{\textbf{User Instruction:} \{instruction\}} \\

\texttt{\textbf{Note:} The reference comment above was based on a comparison between two images.} \\
\texttt{The edited image you are currently assessing is referred to as ``\{image\_label\}'' in that} \\
\texttt{comment. Use this comment as a reference to help you evaluate the Edited Image more} \\
\texttt{accurately.} \\

\texttt{You are provided with two images:} \\
\texttt{1. The Source Image (First image)} \\
\texttt{2. The Edited Image (Second image)} \\

\texttt{Your task is to evaluate the Edited Image against the Source Image and the User} \\
\texttt{Instruction. To do this, you must first assess the image on four critical aspects,} \\
\texttt{provide justifications and absolute scores in 1--4 scale.} \\
\texttt{About the scores: you should try to give \textbf{float scores}. For example, float values are} \\
\texttt{important to reflect fine-grained preferences when you compare two edited images.} \\

\texttt{\textbf{\#\#\# Critical Aspects \& Scoring Rubric}} \\

\texttt{\textbf{1. Text Faithfulness} (How accurately does the output follow the instruction?)} \\
\texttt{- \textbf{4 (Full match):} All key elements (objects, colors, actions) are represented} \\
\texttt{~~exactly as described. No hallucinations or unrequested changes.} \\
\texttt{- \textbf{3 (Minor mismatch):} Most key elements are present, but minor details are} \\
\texttt{~~missing, incorrect, or slightly inaccurate.} \\
\texttt{- \textbf{2 (Some mismatch):} Some key elements are missing, altered, or interpreted} \\
\texttt{~~incorrectly.} \\
\texttt{- \textbf{1 (Major deviations):} Key elements are completely missing, altered, or} \\
\texttt{~~contradicted. Instruction is ignored.} \\

\texttt{\textbf{2. Image Faithfulness} (How well are the non-edited parts and key input} \\
\texttt{elements preserved?)} \\
\texttt{- \textbf{4 (Uses input fully):} All relevant elements from the input (background, style,} \\
\texttt{~~lighting, identity) are accurately preserved or transformed as instructed.} \\
\texttt{- \textbf{3 (Minor mismatch):} Most relevant elements are preserved, but a few aspects} \\
\texttt{~~(e.g., background details, lighting consistency) are missing or incorrectly handled.} \\
\texttt{- \textbf{2 (Partial mismatch):} Some elements are carried over, but key aspects of the} \\
\texttt{~~original image are lost or distorted.} \\
\texttt{- \textbf{1 (Fails to use input):} Key elements of the input image are ignored,} \\
\texttt{~~misinterpreted, or destroyed.} \\

\texttt{\textbf{3. Physical and Visual Quality} (Technical errors, composition, realism, and} \\
\texttt{physics)} \\
\texttt{- \textbf{4 (No noticeable flaws):} The image is physically plausible (correct lighting,} \\
\texttt{~~shadows, geometry, anatomy). No visible artifacts (seams, blurring, noise).} \\
\texttt{- \textbf{3 (Minor flaws):} Small inaccuracies that are noticeable but not strongly} \\
\texttt{~~disruptive (e.g., slight lighting mismatch, minor texture issues).} \\
\texttt{- \textbf{2 (Some flaws):} Clear physical or visual errors that disrupt the image (e.g.,} \\
\texttt{~~incorrect perspective, ``floating'' objects, wrong shadow direction, obvious seams).} \\
\texttt{- \textbf{1 (Severe flaws):} Major physical/visual errors (e.g., impossible geometry,} \\
\texttt{~~distorted anatomy, garbled objects, severe artifacts).} \\
\end{tcolorbox}

\begin{tcolorbox}[title=Pointwise Projection (Part 2), fonttitle=\bfseries, colback=white, colframe=black, breakable]
\small
\texttt{\textbf{4. Text Rendering} (Only if the instruction involves generating text)} \\
\texttt{- \textbf{4 (Full match):} Text is correct, legible, and integrated well.} \\
\texttt{- \textbf{3 (Mostly match):} Minor misspellings or inconsistent capitalization.} \\
\texttt{- \textbf{2 (Partial match):} Major misspellings or distorted text.} \\
\texttt{- \textbf{1 (Major deviations):} Text is unreadable, severely distorted, or missing.} \\
\texttt{~~(Use N/A if no text generation is required).} \\

\texttt{Here is a relevant comment. The comment compares the edited image (referred to as} \\
\texttt{``\{image\_label\}'') with another edited image:} \\
\texttt{\{reference\_comment\}} \\
\texttt{\textbf{Note:} The relevant comment is a hint for you. You can leverage what is useful in it} \\
\texttt{to generate your response. But you MUST NOT mention that this relevant comment is} \\
\texttt{provided when writing the below justifications. Act as if you assess by your own} \\
\texttt{judgement.} \\

\texttt{Output your evaluation in the following format:} \\
\texttt{[ understanding the user request, and what needs to be considered during image} \\
\texttt{editing ]} \\
\texttt{\textbf{\# Detailed Judgement}} \\
\texttt{\textbf{1. Text Faithfulness:}} \\
\texttt{\textbf{\#\# Score:} [ float score ]} \\
\texttt{\textbf{\#\# Justification:} [Detailed explanation of the score]} \\
\texttt{\textbf{2. Image Faithfulness:}} \\
\texttt{\textbf{\#\# Score:} [ float score ]} \\
\texttt{\textbf{\#\# Justification:} [Detailed explanation of the score]} \\
\texttt{\textbf{3. Physical and Visual Quality:}} \\
\texttt{\textbf{\#\# Score:} [ float score ]} \\
\texttt{\textbf{\#\# Justification:} [Detailed explanation of the score]} \\
\texttt{\textbf{4. Text Rendering:}} \\
\texttt{\textbf{\#\# Score:} [ float score or N/A ]} \\
\texttt{\textbf{\#\# Justification:} [Detailed explanation of the score]} \\
\texttt{\textbf{\# Summary:} [Summary of the evaluation]}
\end{tcolorbox}

\subsection{Generate--Critique--Refine (GCR) Loop Prompts}
\label{app:prompt-gcr}

The GCR loop at test time (Section~2.2, Figure~6) uses the trained RationalRewards model in two stages. First, the \emph{critique prompt} evaluates a single generated image across four dimensions with natural language justification. Then, the model generates a \emph{refinement} including a summary of deficiencies and a revised user prompt.

\begin{tcolorbox}[title=GCR Critique and Refinement Prompt (Part 1), fonttitle=\bfseries, colback=white, colframe=black, breakable]
\small
\texttt{User Instruction: \{instruction\}} \\

\texttt{You are provided with two images:}
\begin{enumerate}[leftmargin=*, noitemsep, topsep=0pt]
    \item \texttt{The Source Image (First image)}
    \item \texttt{The Edited Image (Second image)}
\end{enumerate}

\texttt{Your task is to evaluate the Edited Image against the Source Image and the User Instruction.} \\

\texttt{To do this, you must first assess the image on four critical aspects, provide justifications and absolute scores in 1--4 scale.} \\

\texttt{About the scores: you should try to give \textbf{float scores}. For example, float values are important to reflect fine-grained preferences when you compare two edited images.} \\

\texttt{\textbf{\#\#\# Critical Aspects \& Scoring Rubric}} \\

\texttt{\textbf{1. Text Faithfulness} (How accurately does the output follow the instruction?)} \\
\texttt{- \textbf{4 (Full match):} All key elements (objects, colors, actions) are represented exactly as described. No hallucinations or unrequested changes.} \\
\texttt{- \textbf{3 (Minor mismatch):} Most key elements are present, but minor details are missing, incorrect, or slightly inaccurate.} \\
\texttt{- \textbf{2 (Some mismatch):} Some key elements are missing, altered, or interpreted incorrectly.} \\
\texttt{- \textbf{1 (Major deviations):} Key elements are completely missing, altered, or contradicted. Instruction is ignored.} \\

\texttt{\textbf{2. Image Faithfulness} (How well are the non-edited parts and key input elements preserved?)} \\
\texttt{- \textbf{4 (Uses input fully):} All relevant elements from the input (background, style, lighting, identity) are accurately preserved or transformed as instructed.} \\
\texttt{- \textbf{3 (Minor mismatch):} Most relevant elements are preserved, but a few aspects (e.g., background details, lighting consistency) are missing or incorrectly handled.} \\
\texttt{- \textbf{2 (Partial mismatch):} Some elements are carried over, but key aspects of the original image are lost or distorted.} \\
\texttt{- \textbf{1 (Fails to use input):} Key elements of the input image are ignored, misinterpreted, or destroyed.} \\

\texttt{\textbf{3. Physical and Visual Quality} (Technical errors, composition, realism, and physics)} \\
\texttt{- \textbf{4 (No noticeable flaws):} The image is physically plausible (correct lighting, shadows, geometry, anatomy). No visible artifacts (seams, blurring, noise).} \\
\texttt{- \textbf{3 (Minor flaws):} Small inaccuracies that are noticeable but not strongly disruptive (e.g., slight lighting mismatch, minor texture issues).} \\
\texttt{- \textbf{2 (Some flaws):} Clear physical or visual errors that disrupt the image (e.g., incorrect perspective, ``floating'' objects, wrong shadow direction, obvious seams).} \\
\texttt{- \textbf{1 (Severe flaws):} Major physical/visual errors (e.g., impossible geometry, distorted anatomy, garbled objects, severe artifacts).} \\

\texttt{\textbf{4. Text Rendering} (Only if the instruction involves generating text)} \\
\texttt{- \textbf{4 (Full match):} Text is correct, legible, and integrated well.} \\
\texttt{- \textbf{3 (Mostly match):} Minor misspellings or inconsistent capitalization.} \\
\texttt{- \textbf{2 (Partial match):} Major misspellings or distorted text.} \\
\texttt{- \textbf{1 (Major deviations):} Text is unreadable, severely distorted, or missing. (Use N/A if no text generation is required).}
\end{tcolorbox}

\begin{tcolorbox}[title=GCR Critique and Refinement Prompt (Part 2), fonttitle=\bfseries, colback=white, colframe=black, breakable]
\small
\texttt{Output your evaluation in the following format:} \\
\texttt{[ understanding the user request, and what needs to be considered during image editing ]} \\
\texttt{\textbf{\# Detailed Judgement}} \\
\texttt{\textbf{1. Text Faithfulness:}} \\
\texttt{\textbf{\#\# Score:} [ float score ]} \\
\texttt{\textbf{\#\# Justification:} [Detailed explanation of the score]} \\
\texttt{\textbf{2. Image Faithfulness:}} \\
\texttt{\textbf{\#\# Score:} [ float score ]} \\
\texttt{\textbf{\#\# Justification:} [Detailed explanation of the score]} \\
\texttt{\textbf{3. Physical and Visual Quality:}} \\
\texttt{\textbf{\#\# Score:} [ float score ]} \\
\texttt{\textbf{\#\# Justification:} [Detailed explanation of the score]} \\
\texttt{\textbf{4. Text Rendering:}} \\
\texttt{\textbf{\#\# Score:} [ float score or N/A ]} \\
\texttt{\textbf{\#\# Justification:} [Detailed explanation of the score]} \\
\texttt{\textbf{\# Summary:} [Summary of the evaluation]} \\
\texttt{\textbf{\# User Request Refinement:}} \\
\texttt{\textbf{\#\# Refinement Comments:} [Explanation of why the original instruction needs refinement and what constraints should be added]} \\
\texttt{\textbf{\#\# Refined Request:} [Improved, more specific instruction that addresses identified deficiencies]}
\end{tcolorbox}

\paragraph{GCR Loop Logic.} At test time, RationalRewards generates the full critique and refinement output in a single forward pass. If any dimension score falls below the threshold of 3.0, the refined request is extracted and fed back to the generator for re-generation. If all scores are $\geq 3.0$, the original generation is accepted. In our experiments, we use a single-iteration loop (i.e., at most one refinement per image).

\section{Scoring Rubrics}
\label{app:rubrics}

This appendix provides the detailed scoring rubrics for the four assessment dimensions used in pointwise evaluation, referenced in Section~2.1 of the main text. Scores are on a 1--4 integer scale with float-valued interpolation (e.g., 2.5) permitted for fine-grained assessment.

\subsection{Text Faithfulness}

Evaluates how accurately the generated or edited image follows the text instruction.

\begin{table}[h]
\centering
\small
\begin{tabular}{cp{11cm}}
\toprule
\textbf{Score} & \textbf{Description} \\
\midrule
4 (Full match) & All key elements (objects, colors, actions) are represented exactly as described. No hallucinations or unrequested changes. \\
3 (Minor mismatch) & Most key elements are present, but minor details are missing, incorrect, or slightly inaccurate. \\
2 (Some mismatch) & Some key elements are missing, altered, or interpreted incorrectly. \\
1 (Major deviations) & Key elements are completely missing, altered, or contradicted. Instruction is ignored. \\
\bottomrule
\end{tabular}
\caption{Scoring rubric for Text Faithfulness.}
\label{tab:rubric-text-faithfulness}
\end{table}

\noindent\textit{Note: Float-valued scores (e.g., 2.5) interpolate between adjacent anchor descriptions to reflect fine-grained quality distinctions.}

\subsection{Image Faithfulness (Editing Only)}

Evaluates how well the edited image preserves elements of the source image that should remain unchanged.

\begin{table}[h]
\centering
\small
\begin{tabular}{cp{11cm}}
\toprule
\textbf{Score} & \textbf{Description} \\
\midrule
4 (Uses input fully) & All relevant elements from the input (background, style, lighting, identity) are accurately preserved or transformed as instructed. \\
3 (Minor mismatch) & Most relevant elements are preserved, but a few aspects (e.g., background details, lighting consistency) are missing or incorrectly handled. \\
2 (Partial mismatch) & Some elements are carried over, but key aspects of the original image are lost or distorted. \\
1 (Fails to use input) & Key elements of the input image are ignored, misinterpreted, or destroyed. \\
\bottomrule
\end{tabular}
\caption{Scoring rubric for Image Faithfulness.}
\label{tab:rubric-image-faithfulness}
\end{table}

\noindent\textit{Scored as N/A for text-to-image generation tasks where no source image is provided.}

\subsection{Physical and Visual Quality}

Evaluates the physical plausibility and overall visual quality of the generated image.

\begin{table}[h]
\centering
\small
\begin{tabular}{cp{11cm}}
\toprule
\textbf{Score} & \textbf{Description} \\
\midrule
4 (No noticeable flaws) & The image is physically plausible (correct lighting, shadows, geometry, anatomy). No visible artifacts (seams, blurring, noise). \\
3 (Minor flaws) & Small inaccuracies that are noticeable but not strongly disruptive (e.g., slight lighting mismatch, minor texture issues). \\
2 (Some flaws) & Clear physical or visual errors that disrupt the image (e.g., incorrect perspective, ``floating'' objects, wrong shadow direction, obvious seams). \\
1 (Severe flaws) & Major physical/visual errors (e.g., impossible geometry, distorted anatomy, garbled objects, severe artifacts). \\
\bottomrule
\end{tabular}
\caption{Scoring rubric for Physical and Visual Quality.}
\label{tab:rubric-physical-quality}
\end{table}

\subsection{Text Rendering}

Evaluates the quality and accuracy of any text rendered within the generated image.

\begin{table}[h]
\centering
\small
\begin{tabular}{cp{11cm}}
\toprule
\textbf{Score} & \textbf{Description} \\
\midrule
4 (Full match) & Text is correct, legible, and integrated well into the image. \\
3 (Mostly match) & Minor misspellings or inconsistent capitalization. \\
2 (Partial match) & Major misspellings or distorted text. \\
1 (Major deviations) & Text is unreadable, severely distorted, or missing. \\
\bottomrule
\end{tabular}
\caption{Scoring rubric for Text Rendering.}
\label{tab:rubric-text-rendering}
\end{table}

\noindent\textit{Scored as N/A when the instruction does not require text rendering.}

\section{Implementation Details}
\label{app:implementation}

This appendix provides the training hyperparameters, hardware configuration, and RL algorithm details referenced in Section~3 of the main text.





\subsection{RL Fine-Tuning Setup}
\label{app:rl-setup}

We employ DiffusionNFT~\citep{diffusionnft} for RL-based parameter-space optimization. DiffusionNFT is an online RL framework that operates on the forward diffusion process via flow matching, avoiding the need for likelihood estimation, solver restrictions, or classifier-free guidance (CFG) required by reverse-process approaches such as FlowGRPO~\citep{xu2025scalar, jin2025semantic, wu2025hunyuanvideo, lan2025flux, esser2024scaling}.

\paragraph{Algorithm Overview.} DiffusionNFT~\citep{diffusionnft} frames RL for diffusion models as a supervised contrastive learning problem~\citep{wang2025pixel,wang2025illusion}. At each iteration, the algorithm: (1) samples $K$ images from the current policy for a given prompt; (2) evaluates each image with a reward function; (3) splits images into implicit positive (high-reward) and negative (low-reward) subsets; and (4) updates the model via a contrastive flow-matching loss that pushes the policy toward positive generations and away from negative ones. The key theoretical insight is that the velocity-field difference between positive and negative policies defines a \emph{reinforcement guidance direction} $\Delta$ that guarantees policy improvement.

\paragraph{Integration with RationalRewards.} RationalRewards produces per-dimension scores (Text Faithfulness, Image Faithfulness, Physical Quality, Text Rendering) for each generated image. We aggregate these into a scalar reward for the DiffusionNFT loss via equal-weight averaging of applicable dimensions (excluding N/A dimensions). Specifically, for a generated image $x_0$ given prompt $c$:
\begin{equation}
r(x_0, c) = \frac{1}{|\mathcal{D}_{\text{active}}|} \sum_{d \in \mathcal{D}_{\text{active}}} s_d(x_0, c),
\end{equation}
where $s_d$ is the score for dimension $d$ and $\mathcal{D}_{\text{active}}$ is the set of applicable dimensions (e.g., excluding Image Faithfulness for T2I tasks and Text Rendering when no text generation is required).

Algorithm~\ref{alg:rl} provides pseudocode for the RL fine-tuning procedure.

\begin{algorithm}[t]
\caption{RL Fine-Tuning with RationalRewards via DiffusionNFT}
\label{alg:rl}
\begin{algorithmic}[1]
\REQUIRE Flow model $v_\theta$, reference policy $v_{\text{old}} \leftarrow v_\theta$, RationalRewards model $\mathcal{R}$, prompt dataset $\mathcal{C}$, group size $K$, guidance strength $\beta$, EMA schedule $\{\eta_i\}$, number of iterations $N$
\FOR{iteration $i = 1, \ldots, N$}
    \STATE \textcolor{gray}{\textit{// Phase 1: Online Data Collection}}
    \STATE Sample batch of prompts $\{c_j\}_{j=1}^B$ from $\mathcal{C}$
    \FOR{each prompt $c_j$}
        \STATE Generate $K$ images $\{x_0^{(k)}\}_{k=1}^K$ using current sampling policy $v_{\text{old}}$
        \STATE Compute raw rewards: $r_{\text{raw}}^{(k)} \leftarrow \mathcal{R}(x_0^{(k)}, c_j)$ \hfill \textcolor{gray}{\textit{// Multi-dim scores aggregated via Eq.~(D.1)}}
        \STATE Normalize rewards within group: $r^{(k)} \leftarrow 0.5 + 0.5 \cdot \text{clip}\!\left(\frac{r^{(k)} - \bar{r}}{Z_c}, -1, 1\right)$
        \STATE Store $\{c_j, x_0^{(1:K)}, r^{(1:K)}\}$ in buffer $\mathcal{D}$
    \ENDFOR
    \STATE \textcolor{gray}{\textit{// Phase 2: Policy Optimization (Forward Process)}}
    \FOR{each $(c, x_0, r) \in \mathcal{D}$}
        \STATE Sample timestep $t \sim \mathcal{U}(0, 1)$ and noise $\epsilon \sim \mathcal{N}(0, I)$
        \STATE Compute noisy image: $x_t \leftarrow \alpha_t x_0 + \sigma_t \epsilon$
        \STATE Compute flow-matching target: $v \leftarrow \alpha_t' x_0 + \sigma_t' \epsilon$
        \STATE Compute implicit positive velocity: $v_\theta^+ \leftarrow (1 - \beta) v_{\text{old}}(x_t, c, t) + \beta \cdot v_\theta(x_t, c, t)$
        \STATE Compute implicit negative velocity: $v_\theta^- \leftarrow (1 + \beta) v_{\text{old}}(x_t, c, t) - \beta \cdot v_\theta(x_t, c, t)$
        \STATE Compute loss: $\mathcal{L} \leftarrow r \cdot \|v_\theta^+ - v\|^2 + (1 - r) \cdot \|v_\theta^- - v\|^2$
    \ENDFOR
    \STATE Update $\theta$ via gradient descent on $\mathcal{L}$
    \STATE \textcolor{gray}{\textit{// Phase 3: Soft EMA Update of Sampling Policy}}
    \STATE $\theta_{\text{old}} \leftarrow \eta_i \theta_{\text{old}} + (1 - \eta_i) \theta$
\ENDFOR
\RETURN Fine-tuned model $v_\theta$
\end{algorithmic}
\end{algorithm}

\paragraph{RL Hyperparameters.} We employ Low-Rank Adaptation (LoRA)~\citep{hu2022lora} for parameter-efficient fine-tuning. Experiments are conducted on a distributed system comprising 16 NVIDIA A100-80GB GPUs, with 8 GPUs dedicated to model training and 8 GPUs serving the reward model for online evaluation. Table~\ref{tab:rl-hyperparams} summarizes the key hyperparameters.

\begin{table}[h]
\centering
\small
\begin{tabular}{ll}
\toprule
\textbf{Hyperparameter} & \textbf{Value} \\
\midrule
Resolution & 512 $\times$ 512 \\
Guidance Scale (Flux.1 Kontext Dev) & 2.5 \\
Sampling Steps (Training) & 15 (DPM solver) \\
Sampling Steps (Evaluation) & 20 \\
Noise Level & 0.7 \\
Learning Rate & 2e-4 \\
$\beta$ (guidance strength) & 0.0001 \\
Batch Size (per GPU) & 8 \\
Group Size $K$ & 16 (across 16 process groups) \\
Quality Filtering (mean threshold) & 0.9 \\
Quality Filtering (std threshold) & 0.05 \\
LoRA Rank & \texttt{64} \\
LoRA Alpha & \texttt{128} \\
GPU Configuration & 8 $\times$ A100-80GB (training) + 8 $\times$ A100-80GB (reward) \\
Training Wall-Clock Time & $\sim$16 GPU-hours per generator \\
\bottomrule
\end{tabular}
\caption{Hyperparameters for RL fine-tuning via DiffusionNFT.}
\label{tab:rl-hyperparams}
\end{table}

\paragraph{RL Training Data.} We source the RL training prompts from the EditReward Dataset and HPDv3 dataset by selecting prompts whose initial generations receive below-average rewards (mean score $< 3.0$ from RationalRewards), focusing training on cases where the generator has the most room for improvement~\citep{wang2025vl, wang2025emergent}.

\subsection{GCR Loop Configuration}
\label{app:gcr-config}

At inference time, RationalRewards is served via vLLM with prefix caching and paged attention enabled, achieving a per-image overhead of approximately 0.4 seconds for the full critique-and-refinement pass. The refinement threshold is set to 3.0: if any dimension score falls below this value, the refined prompt is used for re-generation. This threshold was selected as the midpoint of the 1--4 scoring scale, corresponding to the boundary between ``minor issues'' (score 3) and ``notable deficiencies'' (score 2) in our rubrics (Appendix~\ref{app:rubrics}).

\section{Dataset and Benchmark Details}
\label{app:data}

\subsection{Training Data Statistics}
\label{app:data-stats}

Table~\ref{tab:data-stats} provides detailed statistics for the training data used in the PARROT pipeline (Section~3, main text).

\begin{table}[h]
\centering
\small
\begin{tabular}{llcccc}
\toprule
\textbf{Source Dataset} & \textbf{Task} & \textbf{Raw Pairs} & \textbf{Post-Filtering Pairs}   & \textbf{Final Pointwise Samples} \\
\midrule
EditReward & Image Editing & 30K & $\sim$21.6K &   $\sim$43.2K \\
HPDv3 & Text-to-Image & \multirow{2}{*}{50K} & \multirow{2}{*}{$\sim$36K}  & \multirow{2}{*}{$\sim$55K} \\
RapidData & Text-to-Image & & & & \\
\bottomrule
\end{tabular}
\caption{Training data composition before and after consistency filtering. Each pairwise sample yields two pointwise projection samples (one per image).}
\label{tab:data-stats}
\end{table}

\noindent We note that our total training scale ($\sim$80K raw pairs, $\sim$57.6K after filtering) is substantially smaller than comparable baselines: EditReward uses 200K pairs and UnifiedReward uses over 1M pairs. Part of this data efficiency stems from the teacher model's pre-trained knowledge, which PARROT distills through structured rationales rather than raw labels.

\subsection{Consistency Filtering Analysis}
\label{app:consistency-analysis}

The consistency filtering step (Phase 2, Section~2.1) retains approximately 72\% of generated rationales overall. We observe the following common failure modes in rejected rationales:

\begin{enumerate}[leftmargin=*]
    \item \textbf{Visual hallucination}: The teacher generates a rationale describing visual content not present in the images (e.g., ``Image A contains a clear sunset in the background'' when no sunset is visible), leading to an incorrect preference prediction when the label hint is removed.
    \item \textbf{Label-ignoring rationales}: Despite the preference anchor, the teacher occasionally generates a rationale that favors the non-preferred image, particularly when the quality difference between images is subtle.
    \item \textbf{Vague, non-predictive reasoning}: The rationale provides generic praise or criticism (e.g., ``Both images are of reasonable quality'') without sufficient discriminative detail to distinguish between the two options.
\end{enumerate}

\subsection{Evaluation Benchmark Summary}
\label{app:benchmarks}

Table~\ref{tab:benchmarks} summarizes all evaluation benchmarks used in this work.

\begin{table}[h]
\centering
\small
\resizebox{\textwidth}{!}{
\begin{tabular}{llccll}
\toprule
\textbf{Benchmark} & \textbf{Task} & \textbf{\# Samples} & \textbf{Evaluation Protocol} & \textbf{Metrics} & \textbf{Reference} \\
\midrule
MMRB2 (T2I) & Preference Prediction & \texttt{1000} & Pairwise comparison & Accuracy & \citet{mmrb2} \\
MMRB2 (Edit) & Preference Prediction & \texttt{1000} & Pairwise comparison & Accuracy & \citet{mmrb2} \\
EditReward Bench & Preference Prediction & \texttt{133} & Pairwise comparison & Accuracy & \citet{editreward} \\
GenAI-Bench (T2I) & Preference Prediction & \texttt{1700} & Pairwise comparison & Accuracy & \citet{jiang2024genai} \\
GenAI-Bench (Edit) & Preference Prediction & \texttt{900} & Pairwise comparison & Accuracy & \citet{jiang2024genai} \\
\bottomrule
\end{tabular}
}
\caption{Summary of evaluation benchmarks.}
\label{tab:benchmarks}
\end{table}

\section{Limitations and Broader Impact}
\label{app:limitations}

\subsection{Limitations}

We acknowledge the following limitations of this work:

\begin{enumerate}[leftmargin=*]
    \item \textbf{Teacher Model Dependence.} The quality of RationalRewards is upper-bounded by the teacher model (Qwen3-VL-32B-Instruct) used to generate training rationales. In domains where the teacher exhibits systematic blind spots---such as fine-grained physics simulation, culturally specific aesthetics, or specialized technical content---the student model inherits these limitations. Future work could explore ensembling multiple teacher models or incorporating human-in-the-loop corrections for high-stakes domains.

    \item \textbf{Bias Inheritance.} Preference datasets (EditReward, HPDv3, RapidData) encode the aesthetic preferences and cultural assumptions of their annotators. The teacher VLM introduces additional biases from its own pretraining data. RationalRewards may therefore systematically favor certain visual styles, demographics, or content types. We have not conducted a comprehensive bias audit, and we encourage users to evaluate the model's behavior on diverse and potentially underrepresented content before deployment.

    \item \textbf{Latent Capability Hypothesis.} Our finding that test-time prompt tuning matches or exceeds RL-based fine-tuning (Section~3.2) supports the hypothesis that generators harbor latent capabilities under-elicited by suboptimal prompts. However, this remains a working hypothesis: we have not validated it at the representation level (e.g., by probing internal activations), and alternative explanations---such as the prompt refinement simply providing additional context that any model would benefit from---cannot be ruled out.

    \item \textbf{Threshold Sensitivity.} The GCR loop uses a fixed threshold of 3.0 to trigger refinement. While this corresponds to a natural boundary in our scoring rubric (Appendix~\ref{app:rubrics}), we have not conducted a comprehensive sensitivity analysis across all benchmarks and generators. The optimal threshold may vary by generator capability and task difficulty.

    \item \textbf{Language and Domain Scope.} All evaluation in this work is conducted on English-language benchmarks. The transferability of RationalRewards' structured critiques to other languages, as well as to non-photorealistic domains (e.g., 3D rendering, video generation, scientific visualization), remains untested.

\end{enumerate}

\subsection{Broader Impact}

RationalRewards and the PARROT framework contribute to the growing ecosystem of tools for evaluating and improving visual generation. We anticipate both positive and negative societal implications:

\paragraph{Positive impacts.}
\begin{itemize}[leftmargin=*]
    \item \textbf{Democratized evaluation:} By providing an open-source, reasoning-based reward model competitive with commercial alternatives, we lower the barrier for researchers and practitioners to evaluate visual generation quality without relying on costly proprietary APIs.
    \item \textbf{Interpretability:} Structured, multi-dimensional critiques provide transparent explanations for quality assessments, enabling users and developers to understand and address specific failure modes rather than optimizing against opaque scalar scores.
    \item \textbf{Accessibility:} The GCR loop can help users with limited prompt engineering experience achieve higher-quality generations by automatically identifying and addressing deficiencies in their instructions.
\end{itemize}

\paragraph{Negative impacts and mitigations.}
\begin{itemize}[leftmargin=*]
    \item \textbf{Misuse potential:} Improved image generation quality could be leveraged for creating misleading visual content, deepfakes, or other harmful media. We note that RationalRewards itself does not generate images but evaluates and critiques them; however, its use as an RL reward or prompt optimizer could amplify generator capabilities.
    \item \textbf{Bias amplification:} As discussed in the limitations, reward models trained on biased preference data may systematically favor certain content types, potentially amplifying existing disparities in visual representation.
    \item We encourage responsible use and recommend that practitioners conduct domain-specific evaluations before deploying RationalRewards in production systems, particularly in sensitive applications.
\end{itemize}